%% file: main_with_authors.tex
\documentclass[10pt,twocolumn,letterpaper]{article}

\usepackage[pagenumbers]{cvpr}      %

\input{preamble}

\usepackage{colortbl}
\usepackage{xcolor}
\usepackage[export]{adjustbox}

\input{math_commands.tex}

\usepackage{amsmath,amsfonts,bm,amssymb,mathtools,amsthm}

\usepackage{algorithm}
\usepackage{algcompatible}
\renewcommand{\COMMENT}[2][.5\linewidth]{%
  \leavevmode\hfill\makebox[#1][l]{\textcolor{gray}{//~#2}}}
  
\usepackage{booktabs}
\usepackage{multirow}
\usepackage{makecell}
\usepackage{pifont}

\newtheorem{remark}[theorem]{Remark}

\definecolor{tabfirst}{rgb}{1, 0.7, 0.7} %
\definecolor{tabsecond}{rgb}{1, 0.85, 0.7} %
\definecolor{tabthird}{rgb}{1, 1, 0.7} %

\definecolor{cvprblue}{rgb}{0.21,0.49,0.74}
\usepackage[pagebackref,breaklinks,colorlinks,citecolor=cvprblue]{hyperref}

\def\paperID{13222} %
\def\confName{CVPR}
\def\confYear{2024}

\title{CoDi: Conditional Diffusion Distillation \\for Higher-Fidelity and Faster Image Generation}

\author{Kangfu Mei\thanks{This work was done during an internship at Google}$^{\,\,\,\,1,2}$,\quad  Mauricio Delbracio$^{1}$,\quad Hossein Talebi$^{1}$,\\ Zhengzhong Tu$^{1}$,\quad 
Vishal M. Patel$^{2}$,\quad Peyman Milanfar$^{1}$\\[.5em]
$^{1}$ Google Research, $^{2}$ Johns Hopkins University \\[0.2em]
\href{https://fast-codi.github.io}{https://fast-codi.github.io}
}

\begin{document}

\twocolumn[{
\maketitle
\vspace{-2.5\baselineskip}
\begin{center}
\includegraphics[width=\linewidth]{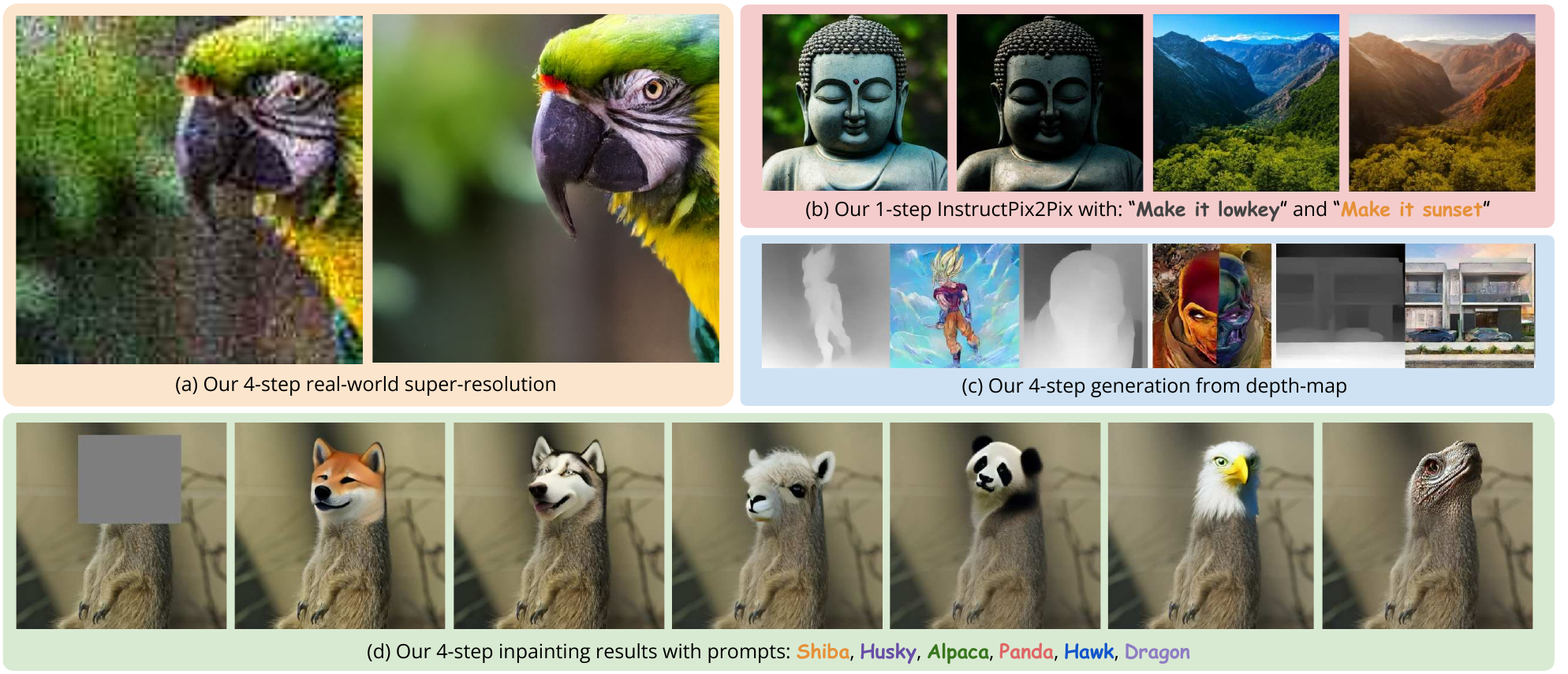}
\vspace{-1.5\baselineskip}
\captionof{figure}{
Our proposed CoDi efficiently distills a conditional diffusion model from an unconditional one, enabling rapid generation of high-quality images under various conditional settings. We demonstrate CoDi's capabilities through generated results across various tasks.
}
\label{fig:teaser}
\end{center}
\label{fig:instance2}
}]

\begin{abstract}
Large generative diffusion models have revolutionized text-to-image generation and offer immense potential for conditional generation tasks such as image enhancement, restoration, editing, and compositing. However, their widespread adoption is hindered by the high computational cost, which limits their real-time application. To address this challenge, we introduce a novel method dubbed CoDi, that adapts a pre-trained latent diffusion model to accept additional image conditioning inputs while significantly reducing the sampling steps required to achieve high-quality results. Our method can leverage architectures such as ControlNet to incorporate conditioning inputs without compromising the model's prior knowledge gained during large scale pre-training. Additionally, a conditional consistency loss enforces consistent predictions across diffusion steps, effectively compelling the model to generate high-quality images with conditions in a few steps. Our conditional-task learning and distillation approach outperforms previous distillation methods, achieving a new state-of-the-art in producing high-quality images with very few steps (e.g., 1-4) across multiple tasks, including super-resolution, text-guided image editing, and depth-to-image generation.
\end{abstract}

\section{Introduction}

Text-to-image diffusion models~\citep{saharia2022photorealistic, rombach2022high, ramesh2022hierarchical} trained on large-scale data~\citep{lin2014microsoft, schuhmann2022laion} have significantly dominated generative tasks by delivering impressive high-quality and diverse results.
A newly emerging trend is to use the prior of pre-trained text-to-image models such latent diffusion models (LDMs)~\cite{rombach2022high} to guide the generated results with external image conditions for image-to-image transformation tasks such as image manipulation, enhancement, or super-resolution~\citep{meng2021sdedit, zhang2023adding}.
Among these transformation processes, the diffusion prior introduced by pre-trained models is shown to be capable of greatly promoting the visual quality of the conditional image generation results~\citep{ruiz2023dreambooth, brooks2023instructpix2pix, liu2023zero, parmar2023zero}.

However, diffusion models heavily rely on an iterative refinement process~\citep{song2020score,saharia2022image,saharia2022palette,whang2022deblurring, delbracio2023inversion} that often demands a substantial number of iterations, which can be challenging to accomplish efficiently.
Their reliance on the number of iterations further increases for high-resolution image synthesis.
For instance, in state-of-the-art text-to-image latent diffusion models~\citep{rombach2022high}, achieving optimal visual quality typically requires $20\mathrm{-}200$ sampling steps (function evaluations), even with advanced sampling methods~\citep{lu2022dpm, karras2022elucidating}.
The slow sampling time significantly impedes practical applications of the aforementioned conditional diffusion models.

Recent efforts to accelerate diffusion sampling predominantly employ distillation methods~\citep{luhman2021knowledge, salimans2022progressive, song2023consistency}.
These methods achieve significantly faster sampling, completing the process in just $4\mathrm{-}8$ steps, with only a marginal decrease in generative performance.
Very recent works~\citep{meng2023distillation, li2023snapfusion} show that these strategies are even applicable for distilling pre-trained large-scale text-to-image diffusion models.

A very common application scenario is to incorporate new conditions into these distilled diffusion models, such as using low-resolution images for super-resoltion~\cite{saharia2022image}, or instruction-tuning for image editing~\cite{brooks2023instructpix2pix}, where the most straightforward way is to directly finetune the distilled text-to-image pre-trained model with new conditional data.
An alternative common approach~\cite{meng2023distillation} is to first finetune the diffusion model with the new conditional data, then conducting distillation on the already-finetuned conditional model.
While these two methods have been demonstrated to accelerate sampling, each has distinct disadvantages in terms of result quality and cross-task flexibility, as discussed below.

In this paper, we introduce a new algorithm for \textbf{\underline{Co}}nditional \textbf{\underline{Di}}stillation which we call \textbf{CoDi} for efficiently adding new controls into distilled models.
Unlike previous distillation methods that rely on finetuning, our method directly distills a diffusion model from a text-to-image pretraining (\eg, StableDiffusion) and ends with a fully distilled conditional diffusion model.
As depicted in Figure~\ref{fig:teaser}, our distilled model is capable of predicting high-quality results in just $1 - 4$ sampling steps.

By design, our method eliminates the need for the original text-to-image data~\cite{schuhmann2022laion, schuhmann2021laion}, a requirement in previous distillation methods (\ie, those that first distill the unconditional text-to-image model), thereby making our method more practical. 
Additionally, our formulation avoids sacrificing the diffusion prior in the pre-trained model during finetuning, a common drawback in the first stage of the finetuning-first procedure.
Our extensive experiments show that our CoDi outperforms previous distillation methods in both visual quality and quantitative metrics, particularly when operating under the same sampling time.

Parameter-efficient distillation methods are a relatively understudied area.
We demonstrate that our method also enables a new \textbf{\underline{P}}arameter-\textbf{\underline{E}}fficient distillation paradigm (\textbf{PE-CoDi}).
It can transform an unconditional diffusion model to conditional tasks by incorporating a small number of additional learnable parameters.
Specifically, our formulation allows for integration with various existing parameter-efficient tuning algorithms, \eg,  ControlNet~\citep{zhang2023adding}.
We show that our distillation process that integrates the ControlNet adapter can efficiently preserve the generative prior in pretraining while adapting the model to new conditioned data.
This new paradigm significantly improves the practicality of different conditional tasks.

Our contributions are summarized as follows:
\begin{itemize}[noitemsep]
\item We propose a new method for image and image-text conditioned generation. It can derive a conditional diffusion model from pretrained text-to-image LDMs for generating high-quality results in only a few sampling steps.
\item The proposed method's efficiency and effectiveness arise from a non-trivial consistency between the model's predictions at different time steps. Enforcing this consistency through learning enables the simultaneous reduction of required sampling steps and the integration of new conditions into the model.
\item  We introduce the first parameter-efficient distillation mechanism that can produce compelling results in just a few steps, while requiring only a small number of additional parameters compared with the pretrained LDMs.
\end{itemize}

\section{Related Work}
\noindent \textbf{Diffusion Distillation.}
To reduce the sampling time of diffusion models, Luhman et al.~\cite{luhman2021knowledge} proposed to learn a single-step student model from the output of the original (teacher) model using multiple sampling steps.
However, this method requires to run the full inference with many sampling steps during training which make it poorly scalable.
Inspired by this, Progressive Distillation~\citep{salimans2022progressive} and its variants, including Guided Distillation~\citep{meng2023distillation} and SnapFusion~\citep{li2023snapfusion}, use a progressive learning scheme for improving the learning efficiency.
A student model learns to predict the output of two steps of the teacher model in one step. Then, the teacher model is replaced by the student model, and the procedure is repeated to progressively distill the mode by halving the number of required steps.
We demonstrate our method by comparing these methods on the conditional generation tasks.
We note that strategies like classifier-free guidance distillation~\cite{meng2023distillation,li2023snapfusion}, or the different adopted sampling techniques~\cite{xu2023restart, zheng2023fast}, are orthogonal to our method, and they could be incorporated in our formulation.
Even though some concurrent works~\cite{xia2023diffir, yue2023resshift} find that tasks like super-resolution requires less sampling steps, we later show that distilling pre-trained diffusion models can still improve the performance in such restoration tasks.

\noindent \textbf{Consistency Distillation.}
A Consistency Model is a single-step generative approach distilled from a pre-trained diffusion model~\cite{song2023consistency}.
The learning is achieved by enforcing a self-consistency in the predicted signal space.
Based on this idea, following work~\cite{gu2023boot,song2023improved, kim2023consistency,lu2023cm} have focus on improving the training techniques.
However, learning consistency models for conditional generation has yet to be thoroughly studied.
In this paper, we compare our method against a baseline approach that enforces self-consistency in an already fine-tuned conditional diffusion model. Our results demonstrate that our conditional distilled model outperforms the baseline approach, indicating the effectiveness of our proposed distillation strategy.

\noindent \textbf{Diffusion Models Adaptations.}
Leveraging the knowledge of pre-trained models for new tasks, known as model adaptation, has gained significant traction in NLP and computer vision domains. This approach utilizes model adapters~\cite{houlsby2019parameter, stickland2019bert, rosenfeld2018incremental, rebuffi2018efficient} and HyperNetworks~\cite{alaluf2022hyperstyle, dinh2022hyperinverter} to effectively adapt pre-trained models to new domains and tasks.
In the context of diffusion models, model adapters have been successfully employed to incorporate new conditions into pre-trained models~\cite{zhang2023adding, mou2023t2i}. Our proposed method draws inspiration from these approaches and introduces a novel application of model adapters: distilling the sampling steps of diffusion models.
Compared to fine-tuning the entire model~\cite{salimans2022progressive}, our method offers enhanced efficiency and flexibility. It enables the adaptation of multiple tasks using the same backbone model.

\section{Background}
\noindent \textbf{Continuous-time VP diffusion model.} A continuous-time variance-preserving (VP) diffusion model~\citep{sohl2015deep, ho2020denoising} is a special case of diffusion models\footnote{What we discussed based on the variance preserving (VP) form of SDE~\citep{song2020score} is equivalent to most general diffusion models like Denoising Diffusion Probabilistic Models (DDPM)~\citep{ho2020denoising}.}.
It has latent variables $\{\bz_t | t \in [0, T]\}$ specified by a noise schedule comprising differentiable functions $\{\alpha_t, \sigma_t\}$ with $\sigma^2_t = 1 - \alpha^2_t$.
The clean data $\bx \sim p_\mathrm{data}$ is progressively perturbed in a (forward) Gaussian process as in the following Markovian structure:
\begin{align}
    q(\bz_t | \bx)& = \mathcal{N}(\bz_t; \alpha_t \bx, \sigma_t^2 \mathbf{I}), \\
    q(\bz_t | \bz_s)& = \mathcal{N}(\bz_t; \alpha_{t|s}\bz_s, \sigma^2_{t|s}\mathbf{I}),
    \label{eq:forward}
\end{align}
where $0 \leq s < t \leq 1$ and $\alpha^2_{t|s} = \alpha_t / \alpha_s$.
Here the latent $\bz_t$ is sampled from the combination of the clean data and random noise by using the reparameterization trick~\citep{kingma2013auto}, which has $\bz_t = \alpha_t\bx + \sigma_t \epsilon$.

\noindent \textbf{Deterministic sampling.}
The aforementioned diffusion process that starts from $\bz_0 \sim p_{\mathrm{data}}(\bx)$ and ends at $\bz_T \sim \mathcal{N}(0, \mathbf{I})$ can be modeled as the solution of an stochastic differential equation (SDE)~\citep{song2020score}. The SDE is formed by a vector-value function $f(\cdot, \cdot): \mathbb{R}^d \to \mathbb{R}^d$, a scalar function $g(\cdot): \mathbb{R} \to \mathbb{R}$, and the standard Wiener process $\bw$ as:
\begin{equation}
    \mathrm{d} \bz_t = f(\bz_t, t)\mathrm{d}t + g(t) \mathrm{d} \bw.
\end{equation}
The overall idea is that the reverse-time SDE that runs backwards in time, can generate samples of $p_\mathrm{data}$ from the prior distribution $\mathcal{N}(0, \mathbf{I})$. This reverse SDE is given by
\begin{equation}
    \mathrm{d} \bz_t =  [f(\bz_t, t) - g(t)^2 \nabla_{\bz} \log p_t(\bz_t)]\mathrm{d} t + g(t) \mathrm{d} \bar{\bw},
    \label{eq:sde}
\end{equation}
where the $\bar{\bw}$ is a also standard Wiener process in reversed time, and $\nabla_\bz \log p_t(\bz_t)$ is the score of the marginal distribution at time $t$. The score function can be estimated by training a score-based model $s_\theta( \bz_t, t) \approx \nabla_z \log p_t (\bz_t)$ with score-matching~\citep{song2020sliced} or a denoising network $\hat{\bx}_\theta(\bz_t, t)$~\citep{ho2020denoising}:
\begin{equation}
    s_\theta(\bz_t,t) := (\alpha_t \hat{\bx}_\theta(\bz_t,t) - \bz_t) / \sigma^2_t.
    \label{eq:scorem}
\end{equation}
Such backward SDE satisfies a special ordinary differential equation (ODE) that allows deterministic sampling given $\bz_T \sim \mathcal{N}(0, \mathbf{I})$. This is known as the \emph{probability flow} (PF) ODE~\citep{song2020score} and is given by
\begin{equation}
    \mathrm{d} \bz_t = [f(\bz_t, t) - \frac{1}{2}g^2(t) s_\theta(\bz_t,t)] \mathrm{d}t,
    \label{eq:pfode}
\end{equation}
where $f(\bz_t, t) = \frac{\mathrm{d} \log \alpha_t}{\mathrm{d} t} \bz_t$, $g^2(t) = \frac{\mathrm{d} \sigma_t^2}{\mathrm{d} t} - 2\frac{\mathrm{d} \log \alpha_t}{\mathrm{d} t}\sigma^2_t$ with respect to $\{\alpha_t, \sigma_t\}$ and $t$ according to \cite{kingma2021variational}.
This ODE can be solved numerically with diffusion samplers like DDIM~\citep{song2020denoising}, where starting from $\hat{\bz}_T \sim \mathcal{N}(0, \mathbf{I})$, we update for $s=t-\Delta t$:
\begin{equation}
    \hat{\bz}_s := \alpha_s \hat{\bx}_\theta(\hat{\bz}_t, t) + \sigma_s (\hat{\bz}_t - \alpha_t \hat{\bx}_\theta (\hat{\bz}_t, t)) / \sigma_t,
    \label{eq:ddim}
\end{equation}
till we reach $\hat{\bz}_0$.

\noindent \textbf{Diffusion models parametrizations.}
Leaving aside the aforementioned way of parametrizing diffusion models with a denoising network (signal prediction) or a score model (noise prediction~\eqref{eq:scorem}), in this work, we adopt a parameterization that mixes both the score (or noise) and the signal prediction. %
Existing methods include either predicting the noise $\hat{\epsilon}_\theta(\bx_t,t)$ and the signal $\hat{\bx}_\theta(\bz_t,t)$ separately using a single network~\citep{dhariwal2021diffusion}, or predicting a combination of noise and signal by expressing them in a new term, like the velocity model $\hat{\bv}_\theta (\bz_t, t) \approx \alpha_t \epsilon - \sigma_t \bx$~\citep{salimans2022progressive}.
Note that one can derive an estimation of the signal and the noise from the velocity one,
\begin{equation}
    \hat{\bx} = \alpha_t \bz_t - \sigma_t \hat{\bv}_\theta (\bz_t, t), \, \mathrm{and}\,\,\, \hat{\epsilon} = \alpha_t \hat{\bv}_\theta (\bz_t, t)  + \sigma_t \bz_t.
    \label{eq:v}
\end{equation}
Similarly, DDIM update rule (\eqref{eq:ddim}) can be rewritten in terms of the velocity parametrization:
\begin{equation}
        \hat{\bz}_s := \alpha_s (\alpha_t \hat{\bz}_t - \sigma_t \hat{\bv}_\theta (\hat{\bz}_t, t)) + \sigma_s (\alpha_t \hat{\bv}_\theta (\hat{\bz}_t, t)  + \sigma_t \hat{\bz}_t).
    \label{eq:vddim}
\end{equation}

\noindent \textbf{Self-consistency property.}
To accelerate inference, \cite{song2023consistency} introduced the idea of consistency models.
Let $s_\theta(\cdot, t)$ be a pre-trained diffusion model trained on data $\bx \sim \mathcal{O}_{data}$. Then, a consistency function $f_\phi(\bz_t, t)$ should satisfy that~\citep{song2023consistency} where $f_\phi (\bx, 0) = \bx$ and
\begin{equation}
    f_\phi (\bz_t, t) = f_\phi (\bz_{t'}, t'),\,\, \forall t, t' \in [0, T],
    \label{eq:cm}
\end{equation}
where $\{\bz_t\}_{t\in[0, T]}$ is the solution trajectory of the probability flow ODE (PF-ODE) (\eqref{eq:pfode}).
A boundary condition, \ie, $f_\phi (\bx, 0) = \bx$ is parameterized with skip connections for ensuring continuous properties similar as done in previous works ~\cite{karras2022elucidating,balaji2022ediffi,song2023consistency}:
\begin{equation}
    F_\phi (\bz_t, t) = c_\mathrm{skip}(t) \bx + c_\mathrm{out}(t) f_\phi (\bz_t, t),
    \label{eq:boundary}
\end{equation}
where $c_\mathrm{skip}(0)=1$, $c_\mathrm{out}(0)=0$.
In practice, $f_\phi(\bz_t, t)$ is usually a denoising network that is distilled from a pre-trained diffusion model.
We later show that we can replace the frozen PF-ODE with the distillation network and thus fit the PF-ODE for new conditional data during distillation.

\section{Method}

\subsection{From Unconditional to Conditional}
\label{sec:ca}
In order to utilize the image generation prior encapsulated by the pre-trained unconditional\footnote{The discussed unconditional models include text-conditioned image generation models, \emph{e.g.}, StableDiffusion~\citep{rombach2022high} and Imagen~\citep{saharia2022photorealistic}, which are only conditioned on text prompts.} diffusion model, we first propose to adapt the unconditional diffusion model into a conditional version for the conditional data $(\bx, c) \sim p_\mathrm{data}$.
Similar to the zero initialization technique used by controllable generation~\citep{nichol2021improved, zhang2023adding}, our method adapts the unconditional pre-trained architecture by using an additional conditional encoder.

To elaborate, we take the widely used U-Net as the diffusion network. Let us introduce the conditional-module by duplicating the encoder layers of the pretrained network. Then, let $\boldsymbol{h}_\theta(\cdot)$ be the encoder features of the pretrained network, and $\boldsymbol{h}_\eta(\cdot)$ be the features on the additional conditional encoder. We define the new encoder features of the adapted model by
\begin{equation}
    \boldsymbol{h}_\theta(\bz_t)'  = (1 - \mu) \boldsymbol{h}_\theta(\bz_t) + \mu \boldsymbol{h}_\eta(c),
\end{equation}
where $\mu$ is a learnable scalar parameter, initialized to $\mu=0$.
Starting from this zero initialization, we can adapt the unconditional architecture into a conditional one.
Thus, our conditional diffusion model $\hat{\bw}_\theta(\bz_t,c,t)$ is the result of adapting the pre-trained unconditional diffusion model $\hat{\bv}_\theta(\bz_t,t)$ with the conditional features $\boldsymbol{h}_\eta(c)$.

\subsection{A New Conditional Diffusion Consistency}
\label{sec:method}
Our core idea is to optimize the adapted conditional diffusion model $\hat{\bw}_\theta(\bz_t, c, t)$ from $\hat{\bv}_\theta(\bz_t, t)$, so it satisfies a conditional diffusion consistency property:
\begin{equation}
    \hat{\bw}_\theta(\bz_t, c, t) = \hat{\bw}_\theta(\hat{\bz}_s, c, s),\,\, \forall t, s \in [0, T],
    \label{eq:cdc}
\end{equation}
where the $\hat{\bz}_s$ belong to the probability flow ODE (\eqref{eq:pfode}) of the adapted model. Note that this consistency property differs from the one in consistency models~\cite{song2023consistency} in the probability flow ODE model used for sampling $\hat{\bz}_s$ and the consistency loss space.
To motivate this formulation, let us introduce the following general remark.

\begin{remark}
\label{remark:consistency}
If a diffusion model, parameterized by $\hat{\bv}_\theta(\bz_t, t)$, satisfies the self-consistency property (\eqref{eq:cm}) on the noise prediction $\hat{\epsilon}_\theta(\bz_t, t) = \alpha_t \hat{\bv}_\theta(\bz_t, t) + \sigma_t \bz_t$, then it also satisfies the self-consistency property on the signal prediction  $\hat{\bx}_\theta(\bz_t,  t) = \alpha_t \bz_t - \sigma_t \hat{\bv}_\theta(\bz_t,  t)$.
\end{remark}
The proof is a direct consequence of change of variables from noise into signal and is given in Appendix.
Based on this general remark, we claim that we can optimize the conditional diffusion model $\hat{\bw}_\theta(\bz_t, c, t)$  to jointly learn to enforce the self-consistency property on the noise prediction $\hat{\epsilon}_\theta(\bz_t, c, t)$ and the new conditional generation  $(\bx, c) \sim p_\mathrm{data}$ with the signal prediction $\hat{\bx}_\theta(\bz_t, c, t)$.
We then impose the boundary condition for consistency distillation by parameterizing the noise prediction $\hat{\epsilon}_\theta(\bz_t, c, t)$ with the same skip connections of \eqref{eq:boundary}.

\noindent \textbf{Prediction of $\hat{\bz}_s$.}
In the distillation process given by~\eqref{eq:loss}, the latent variable $\hat{\bz}_s$ is achieved by running one step of a numerical ODE solver.
Consistency models~\citep{song2023consistency} solve the ODE using the Euler solver, while progressive distillation~\citep{salimans2022progressive} and guided distillation~\citep{meng2023distillation} run two steps using the DDIM sampler (\eqref{eq:ddim}).

We propose an alternative prediction for $\hat{\bz}_s$ that leverages the adapted diffusion model, $\hat{\bx}_\theta(\bz_t, c, t)$, as opposed to the conventional frozen pretraining one.
We then sample $\Hat{\bz}_s$ in the adapted diffusion model PF-ODE by
\begin{equation}
\hat{\bz}_s = \alpha_s \hat{\bx}_\theta(\bz_t, c, t) + \sigma_s \epsilon,
\, \text{with}\,\,
\bz_t = \alpha_t \bx + \sigma_t \epsilon, 
\label{eq:prevs}
\end{equation}
and $\epsilon \sim \mathcal{N}(0, \mathbf{I})$.
This novel formulation effectively harmonizes the conflicting optimization directions between consistency distillation from pretrained data and conditional guidance from conditional data.

\input{tables/regularization}

\noindent \textbf{Training scheme.}
Inspired by consistency models~\citep{song2023consistency}, we use the exponential moving averaged parameters $\theta^-$ as the {target network} for stabilize training.
Then, we can minimize the following training loss for conditional distillation:

\begin{equation}
    \small
    \label{eq:loss}
    \mathcal{L}(\theta)\!:=\!\mathbb{E}[ {d_\epsilon}(\hat{\epsilon}_{\theta^{\mbox{-}}}\!(\hat{\bz}_s,\!s,\!c), \hat{\epsilon}_\theta (\bz_t,\!t,\!c))) + d_\bx(\bx, \hat{\bx}_\theta( \bz_t, t, c)].
\end{equation}

\noindent where $d_\epsilon(\cdot, \cdot)$ and $d_\bx(\cdot, \cdot)$ are two distance functions to measure difference in the noise space and in the signal space respectively. Note that the total loss is a balance between the conditional guidance given by $d_\bx$, and the noise self-consistency property given by $d_\epsilon$. 

The overall conditional distillation algorithm is presented in Appendix.
In the following, we will detail how we sample $\hat{\bz}_s$ and discuss other relevant hyperparameters in our method (e.g.,  $d_\bx$).

\subsection{Effects of Different Conditional Guidance}
\label{sec:cg}
To finetune the adapted diffusion model with the new conditional data, our conditional diffusion distillation loss in \eqref{eq:loss} penalizes the difference between the predicted signal $\hat{\bx}_\theta(\bz_t, c, t)$ and the corresponding image $\bx$ with a distance function $d_\bx(\cdot, \cdot)$ for distillation learning.

Here we investigate the impact of the distance function $d_\bx(\cdot, \cdot)$ in the conditional guidance. According to both qualitative and quantitative results, shown in Figure~\ref{fig:regularization}, different distance functions lead to different behaviours when doing multi-step sampling (inference).
If  $d_\bx = \|\cdot\|^2$   in the pixel space or the encoded space, \emph{i.e.}, $\| \bx - \mathbb{E}(\mathbb{D}(\hat{\bx}_\theta(\bz_t, c, t))) \|^2_2$ and $\| \mathbb{D}(\bx) - \mathbb{D}(\hat{\bx}_\theta(\bz_t, c, t)) \|^2_2$, multi-step sampling leads to more smooth and blurry results. 
If instead we adopt a perceptual distance in the pixel space, \emph{i.e.}, $\mathcal{F}_{\mathrm{lpips}}(\mathbb{D}(\bx), \mathbb{D}(\hat{\bx}_\theta(\bz_t, c, t)))$, the iterative refinement in the multi-step sampling leads to over-saturated results. 
Overall, by default we adopted the $\ell_2$ distance in the latent space since it leads to better visual quality and achieve the optimal FID with 4 sampling steps in Figure~\ref{fig:regularization}.

\subsection{Parameter-Efficient Conditional Distillation}
\label{sec:ped}

\begin{figure}

\centering
\includegraphics[width=\linewidth]{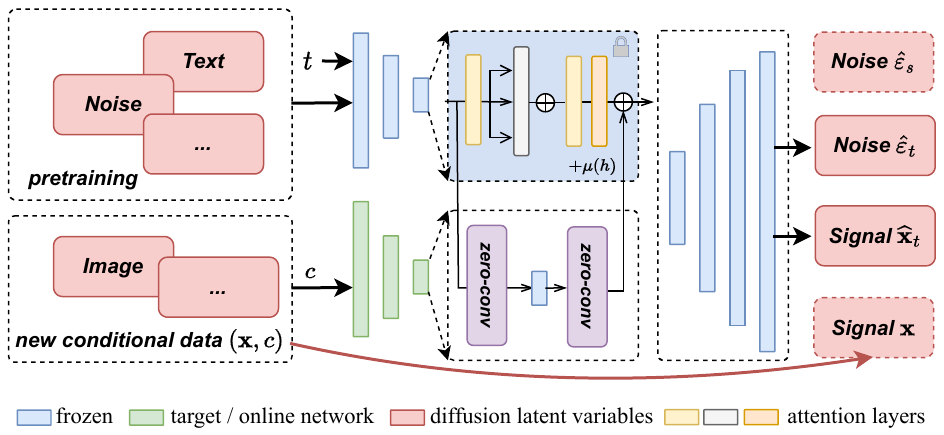}
\caption{\small Network architecture illustration of our parameter-efficient conditional distillation framework.}
\label{fig:controlnet}
\vspace{-1\baselineskip}
\end{figure}

Our method offers the flexibility to selectively update parameters pertinent to distillation and conditional finetuning, leaving the remaining parameters frozen.
This leads us to introduce a new fashion of parameter-efficient conditional distillation, aiming at unifying the distillation process across commonly-used parameter-efficient diffusion model finetuning, including ControlNet~\citep{zhang2023adding}, T2I-Adapter~\citep{mou2023t2i}, etc.
We highlight the ControlNet architecture illustrated in Figure~\ref{fig:controlnet} as an example.
This model duplicates the encoder part of the denoising network, highlighted in the green blocks, as the condition-related parameters.
Our method can then optimizes the conditional guidance and the consistency by only updating the duplicated encoder. 

\begin{table}[h]
    \centering
    \small
    \resizebox{\linewidth}{!}{%
    \begin{tabular}{r|ccccc}
        \toprule
        & CM-I & CM-II & GD-I & GD-II & Ours \\
        \midrule
        stage-1 & distill & finetune & distill & finetune & conditional distill \\
        stage-2 & finetune & distill & finetune & distill & n.a. \\
        \midrule
        \midrule
        & \ding{55} & \ding{51} & \ding{51} & \ding{51} & \ding{51} \\
        \bottomrule
    \end{tabular}
    }
    \vspace{-2pt}
    \caption{We compare previous distillation methods by applying them to a T2I LDMs and then finetuning the distilled models (CM-X), and also distillation methods by directly applying them into the finetuned LDMs (GD-X). Since fine-tuning a distilled consistency model within the existing diffusion loss framework is not feasible, we excluded it from our comparison.}
    \label{tab:baselines}
\end{table}

\begin{figure*}[ht]
    \centering
    \def\xwidth{0.11\linewidth}
    \def\ywidth{0.11\linewidth}
    \setlength{\tabcolsep}{1pt}
    \renewcommand\arraystretch{1.1}
    \begin{tabular}[t]{c c c c c c c c c}\\
     \cellcolor{tabfirst} \includegraphics[width=\xwidth]{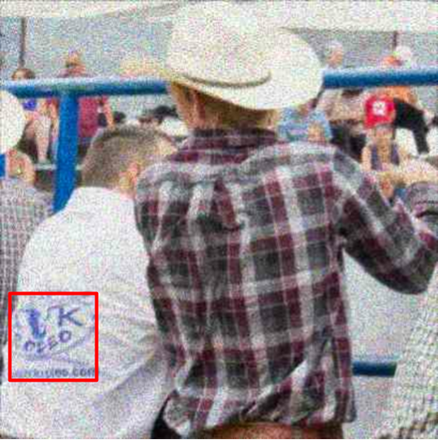} & \cellcolor{tabfirst} \includegraphics[width=\xwidth]{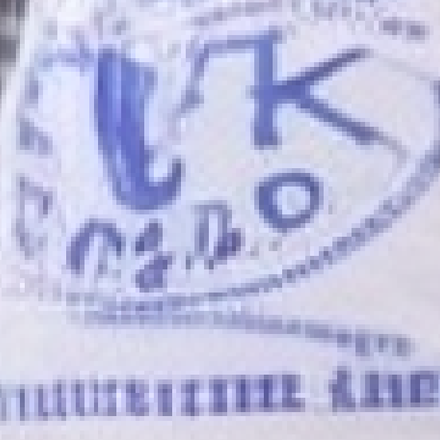} & \cellcolor{tabfirst} \includegraphics[width=\xwidth]{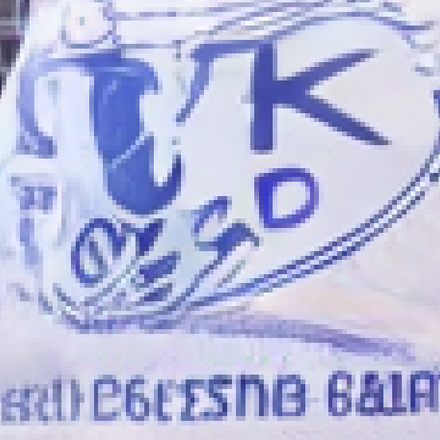} & \cellcolor{tabfirst} \includegraphics[width=\xwidth]{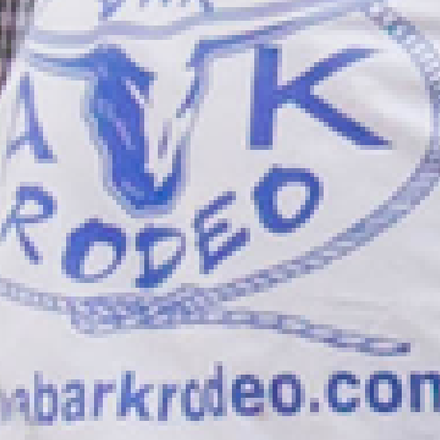} & 
    & \cellcolor{tabsecond} \includegraphics[width=\xwidth]{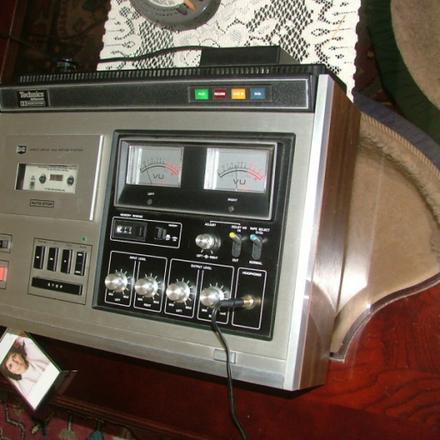} 
    & \cellcolor{tabsecond} \includegraphics[width=\xwidth,height=\ywidth]{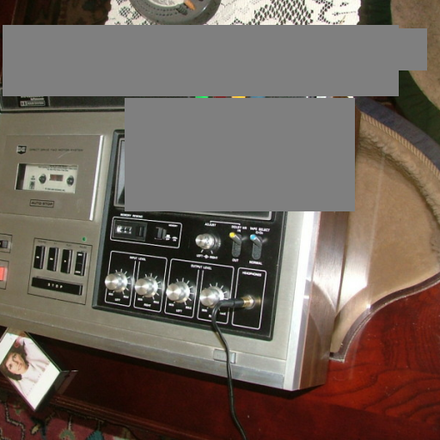}
    & \cellcolor{tabsecond} \includegraphics[width=\xwidth,height=\ywidth]{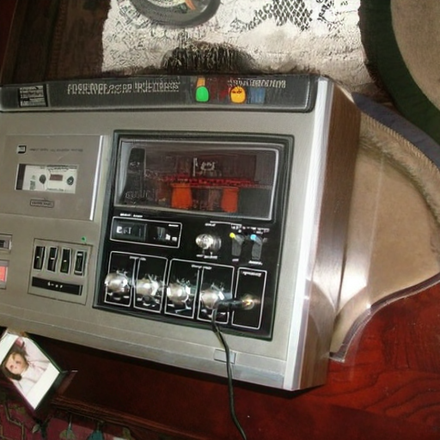}
    & \cellcolor{tabsecond} \includegraphics[width=\xwidth]{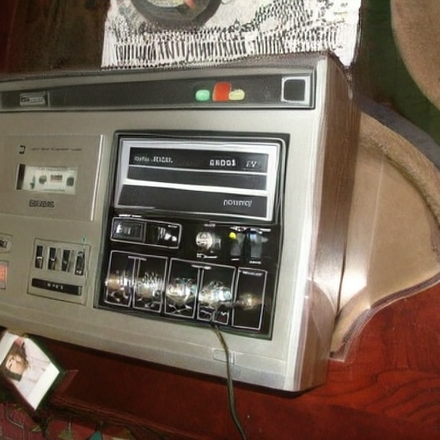}
    \\
    \cellcolor{tabfirst} \includegraphics[width=\xwidth]{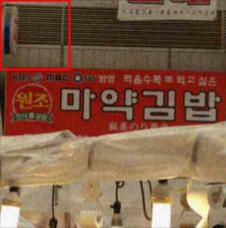} & \cellcolor{tabfirst} \includegraphics[width=\xwidth]{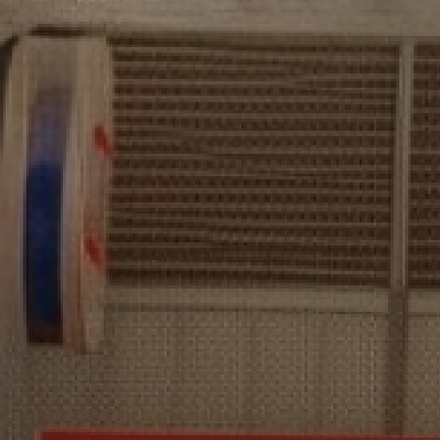} & \cellcolor{tabfirst} \includegraphics[width=\xwidth]{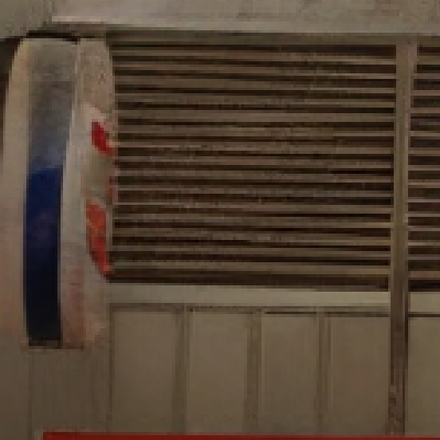} & \cellcolor{tabfirst} \includegraphics[width=\xwidth]{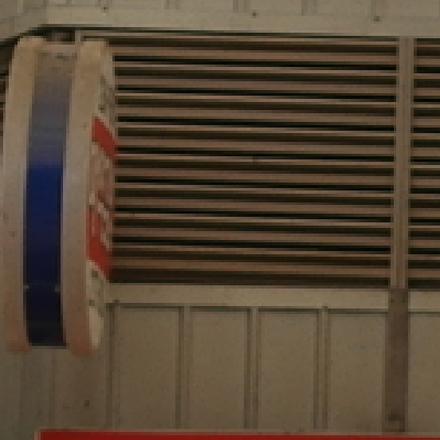} & 
    & \cellcolor{tabsecond} \includegraphics[width=\xwidth]{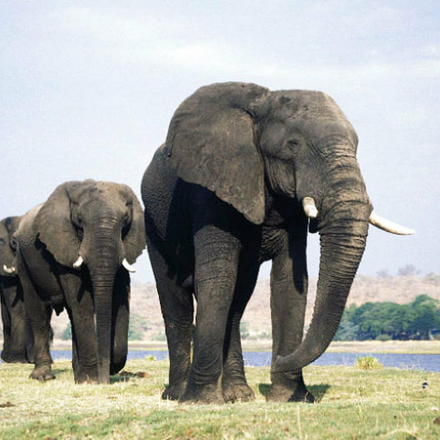} 
    & \cellcolor{tabsecond} \includegraphics[width=\xwidth,height=\ywidth]{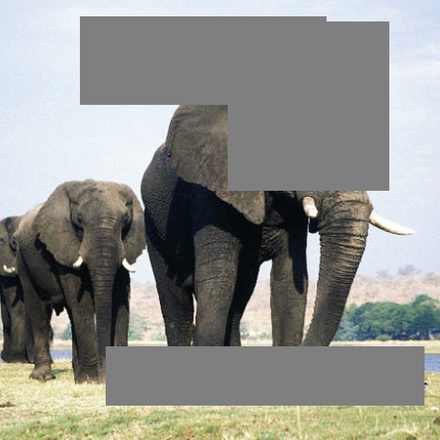}
    & \cellcolor{tabsecond} \includegraphics[width=\xwidth,height=\ywidth]{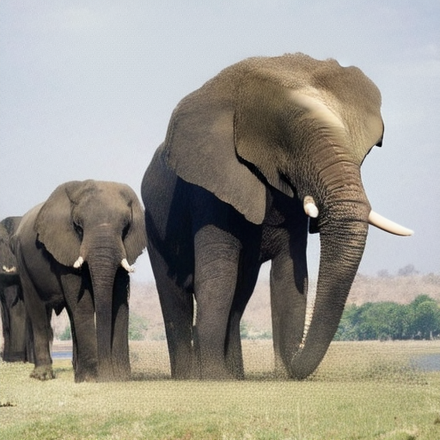}
    & \cellcolor{tabsecond} \includegraphics[width=\xwidth]{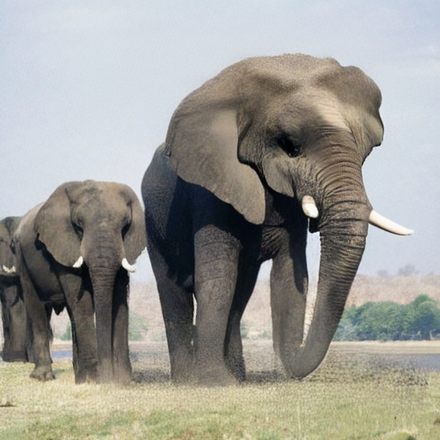}
    \\
    \cellcolor{tabfirst} \includegraphics[width=\xwidth]{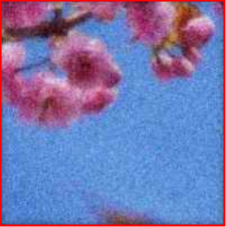} & \cellcolor{tabfirst} \includegraphics[width=\xwidth]{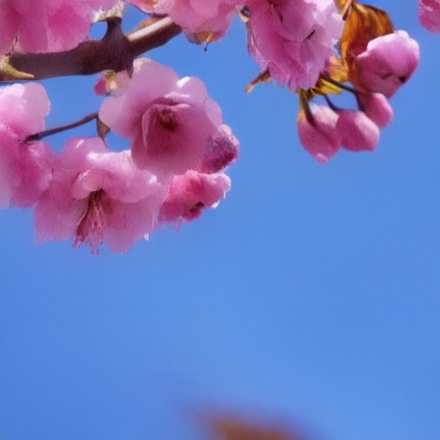}& \cellcolor{tabfirst} \includegraphics[width=\xwidth]{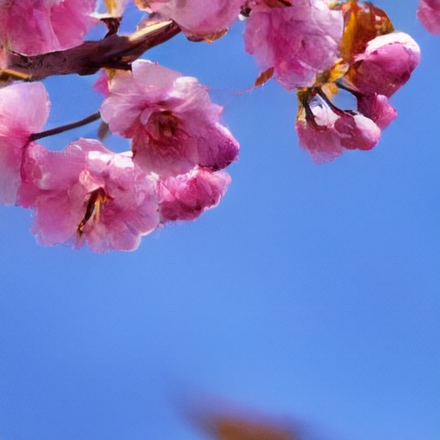} & \cellcolor{tabfirst} \includegraphics[width=\xwidth]{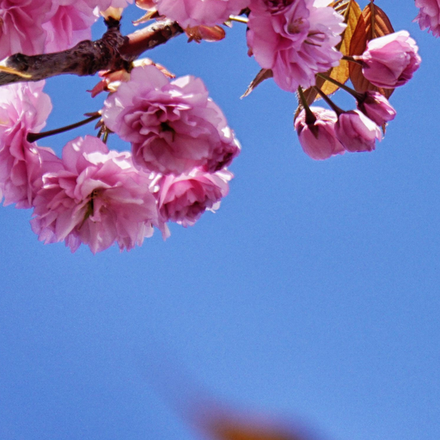} & 
    & \cellcolor{tabsecond} \includegraphics[width=\xwidth]{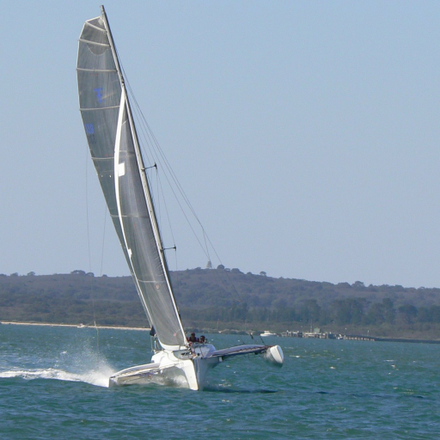} 
    & \cellcolor{tabsecond} \includegraphics[width=\xwidth,height=\ywidth]{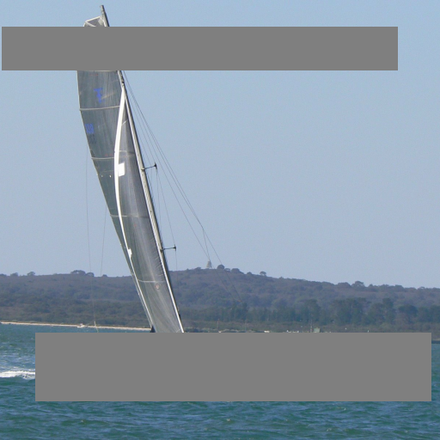}
    & \cellcolor{tabsecond} \includegraphics[width=\xwidth,height=\ywidth]{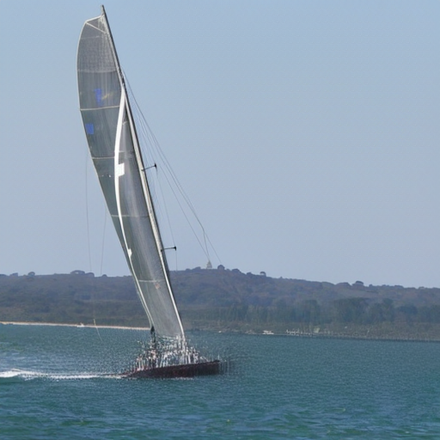}
    & \cellcolor{tabsecond} \includegraphics[width=\xwidth]{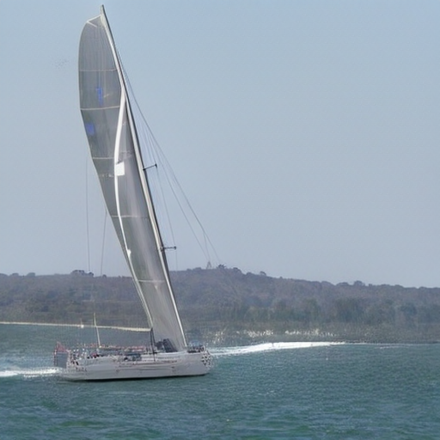} 
    \\
    \cellcolor{tabfirst} \includegraphics[width=\xwidth]{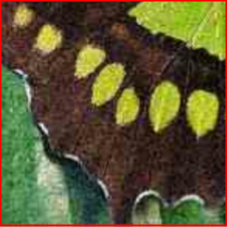} & \cellcolor{tabfirst} \includegraphics[width=\xwidth]{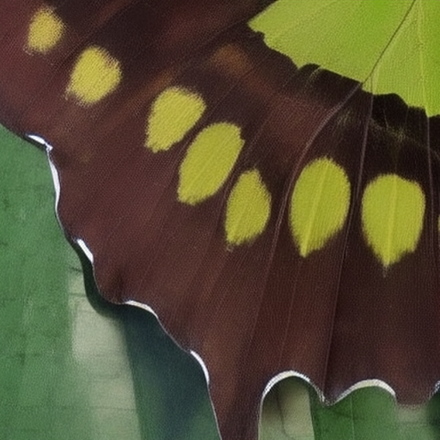} & \cellcolor{tabfirst} \includegraphics[width=\xwidth]{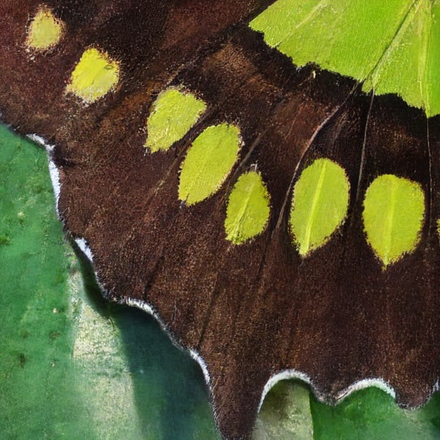} & \cellcolor{tabfirst} \includegraphics[width=\xwidth]{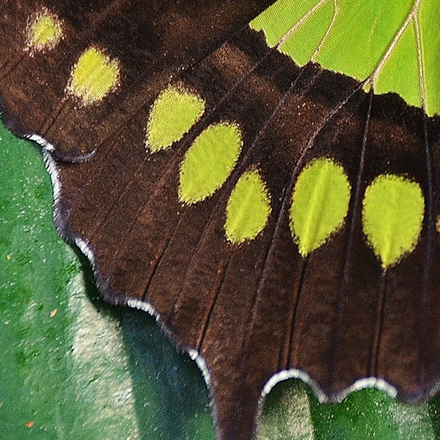} & 
    & \cellcolor{tabsecond} \includegraphics[width=\xwidth]{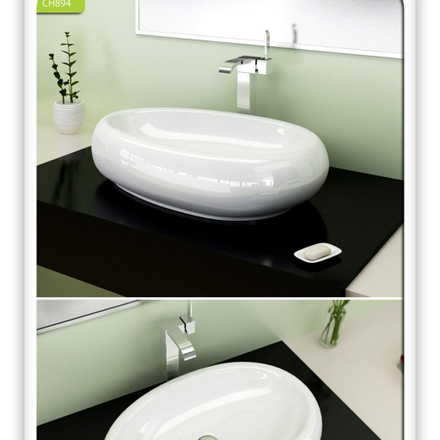} 
    & \cellcolor{tabsecond} \includegraphics[width=\xwidth,height=\ywidth]{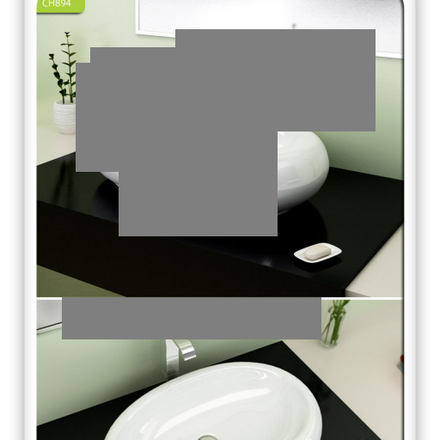}
    & \cellcolor{tabsecond} \includegraphics[width=\xwidth,height=\ywidth]{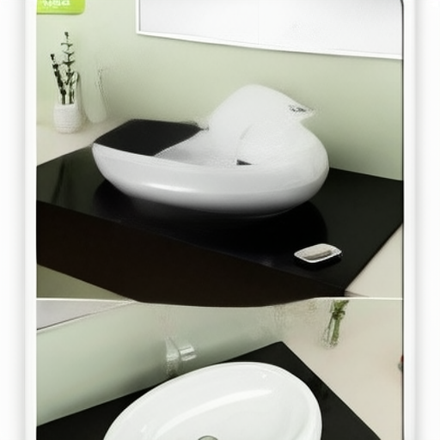}
    & \cellcolor{tabsecond} \includegraphics[width=\xwidth]{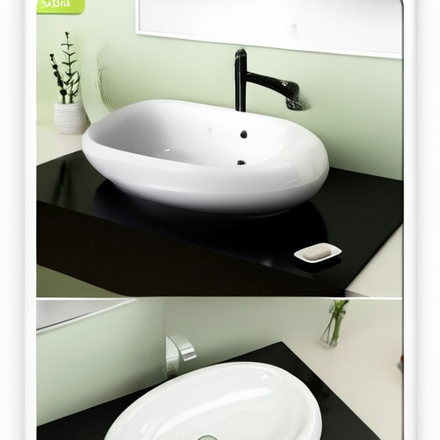}
    \\
    \scriptsize LR & \scriptsize CM-II & \scriptsize CoDi (Ours)  & \scriptsize HR & & \scriptsize Ground Truth & \scriptsize Mask & \scriptsize CM-II & \scriptsize PF-CoDi (Ours) 
    \end{tabular}
    \vspace{-1\baselineskip}
    \caption{We show the results sampled in 4 steps by different models. Samples generated according to the low-resolution images (left) and masks (right) respectively. Please see our supplement for many more examples such as visual comparisons with the other methods.}
    \label{fig:srinpainting}
\end{figure*}

\begin{table*}[ht]
    \centering
    \begin{tabular}{@{}cc@{}}
    \begin{minipage}[t]{.48\linewidth}
    \centering
    \small
    \renewcommand{\arraystretch}{0.89}
    {%
    \begin{tabular}{rrcc}
        \multicolumn{4}{l}{\textbf{Super-resolution (DF2K)}} \\
        \toprule
        Sampling Steps & Methods &  FID $\downarrow$ & LPIPS $\downarrow$ \\
        \midrule
        1 step & RealESRGAN~\cite{wang2022realesrgan} & 37.640 & 0.3112  \\
        200 steps & StableSR~\cite{wang2023exploiting} & 24.440 & 0.3114  \\
        4 steps & DiffIR~\cite{xia2023diffir} & 31.719 & 0.3088 \\
        4 steps & ControlNet~\cite{zhang2023adding}  & 34.56 & 0.3381  \\
        \midrule
        250 steps & LDMs~\cite{rombach2022high} & \cellcolor{tabfirst} 19.200 &  \cellcolor{tabfirst} 0.2639  \\
        50 steps & LDMs~\cite{rombach2022high} & \cellcolor{tabsecond} 19.231 & \cellcolor{tabsecond} 0.2603  \\
        20 steps & LDMs~\cite{rombach2022high} & 20.510 & 0.2627  \\
        8 steps & LDMs~\cite{rombach2022high} & 24.493 & 0.2789  \\
        6 steps & LDMs~\cite{rombach2022high} & 26.338 & 0.2873  \\
        4 steps  & LDMs~\cite{rombach2022high} & 29.266 & 0.3014 \\
        4 steps & + DPM Solve~\cite{lu2022dpm}  & 28.936 & 0.3077  \\
        4 steps & + DPM Solver++~\cite{lu2022dpm2}  & 28.937 & 0.3073  \\
        \midrule
        & GD-I~\cite{meng2023distillation}  & 27.806 & 0.3202 \\
        & GD-II~\cite{meng2023distillation} & 23.675 & 0.2796 \\
        & CM-II (frozen) ~\cite{song2023consistency}  & 28.088 & 0.3192 \\
        & CM-II ~\cite{song2023consistency}  & 27.810 & 0.3172 \\
        \midrule
        4 steps & \textbf{PE-CoDi} (Ours)  & 25.214 & 0.2941 \\
        & \textbf{CoDi} (Ours)  &  \cellcolor{tabthird} 19.637 & \cellcolor{tabthird} 0.2656 \\
        \bottomrule
    \end{tabular}
    }
    \end{minipage}
    &
    \begin{tabular}{@{}c@{}}

    \begin{minipage}[t]{.48\linewidth}
	{
	\centering
    \small
    \renewcommand{\arraystretch}{0.82}
	\begin{tabular}{rrcc}
	    \multicolumn{4}{l}{\textbf{Inpainting (ImageNet)}} \\
        \toprule
        Sampling Steps & Methods &  FID & LPIPS \\
        \midrule
        1000 steps & Palette~\cite{saharia2022palette} & \cellcolor{tabfirst} 13.151 & -  \\
        250 steps & Repaint~\cite{lugmayr2022inpainting} & - & 0.2827 \\
        50 steps & ControlNet~\cite{zhang2023adding} & \cellcolor{tabthird} 14.895 & \cellcolor{tabsecond} 0.2260  \\
        \midrule
        4 steps  & ControlNet~\cite{rombach2022high} & 20.205 & 0.2635 \\
        & + DPM Solver++~\cite{lu2022dpm2} & 19.941 & 0.2644 \\
        & CM-II~\cite{song2023consistency} & 17.710 & 0.2580  \\
        & GD-II~\cite{meng2023distillation} & 15.95 &  \cellcolor{tabthird} 0.2452  \\
        \midrule
        4 steps & \textbf{PE-CoDi} (Ours) &  \cellcolor{tabsecond} 14.700 & \cellcolor{tabfirst} 0.2231  \\
        \bottomrule
	\end{tabular}
	}
	{
	\centering
    \small
    \renewcommand{\arraystretch}{0.8}
	\begin{tabular}{rrcc}
	    \\
        \\
	    \multicolumn{4}{l}{\textbf{Text-guided Depth-to-image (WebLI)}} \\
        \toprule
        Sampling Steps & Methods &  FID & CLIP \\
        \midrule
        250 steps & ControlNet~\cite{zhang2023adding} & \cellcolor{tabfirst} 20.884 & \cellcolor{tabfirst} 0.2910  \\
        \midrule
        4 steps  & ControlNet~\cite{zhang2023adding} & \cellcolor{tabthird}  29.780 & \cellcolor{tabthird} 0.2854 \\
        & + DPM Solver++~\cite{lu2022dpm2} & 32.208 & 0.2834  \\
        & CM-II~\cite{song2023consistency} & 27.640 & 0.2869  \\
        & GD-II~\cite{meng2023distillation} & 26.51 &  \cellcolor{tabthird} 0.2870  \\
        \midrule
        4 steps & \textbf{PE-CoDi} (Ours) & \cellcolor{tabsecond} 23.047 & \cellcolor{tabsecond} 0.2874 \\
        \bottomrule
	\end{tabular}
	}
	\end{minipage}
    \end{tabular}
	\end{tabular}
\vspace{-2pt}
    \caption{Quantitative performance comparisons between the baselines and our methods. Our model can achieve comparable performance in 4 steps than models sampled in 250 steps. The 4-step sampling results of our parameters-efficient distillation (PE-CoDi) is comparable with the original 8-step sampling results, while PE-CoDi doesn't sacrifice the original generative performance with frozen backbone.
    }
    \label{tab:performance}
    \vspace{-1\baselineskip}
\end{table*}

\begin{figure*}[!ht]
    \def\xwidth{0.329\textwidth}
    \begin{subfigure}[b]{\xwidth}
    \centering
    \includegraphics[width=\linewidth]{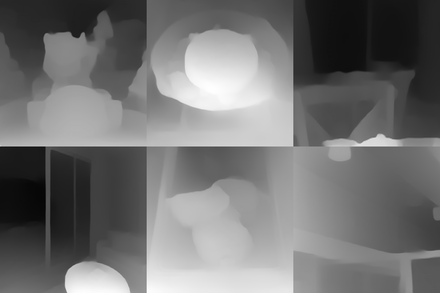}
    \caption{Depth}
    \end{subfigure}
    \hfill
    \begin{subfigure}[b]{\xwidth}
    \centering
    \includegraphics[width=\linewidth]{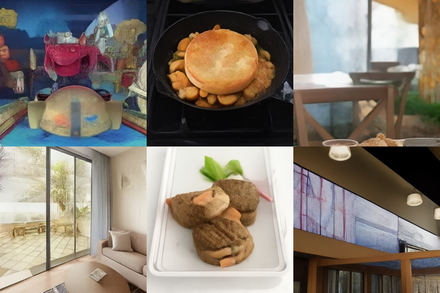}
    \caption{ControlNet}
    \end{subfigure}
    \hfill
    \begin{subfigure}[b]{\xwidth}
    \centering
    \includegraphics[width=\linewidth]{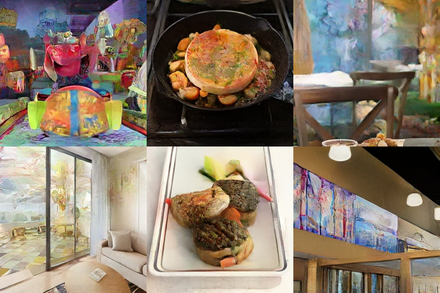}
    \caption{\textbf{CoDi (Ours)}}
    \end{subfigure}

    \vspace{-.5\baselineskip}
    \caption{Samples generated according to the depth image (left) from ControlNet sampled in 4 steps (middle), and ours from the unconditional pretraining sampled in 4 steps (right). Please see our supplement for many more examples.}
    \label{fig:depth}
    \vspace{-.5\baselineskip}
\end{figure*}

\begin{figure*}[!ht]
    \centering
    \setlength{\tabcolsep}{1pt}
    \def\xwidth{0.325\textwidth}
    \def\xxxwidth{0.495\linewidth}
    \begin{minipage}[b]{\xxxwidth}
    \begin{tabular}{c c c}
    \small Input & \small IP2P (200 steps) & \small \textbf{CoDi (Ours)} (1 step) \\
    \includegraphics[width=\xwidth]{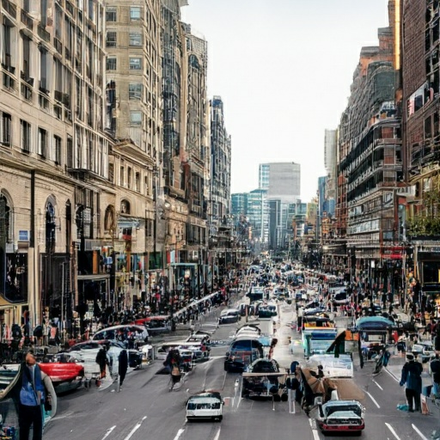} & \includegraphics[width=\xwidth]{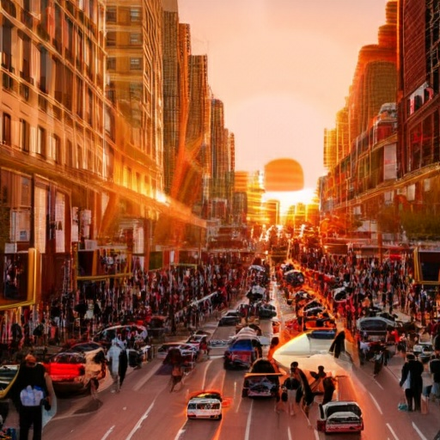} &
    \includegraphics[width=\xwidth]{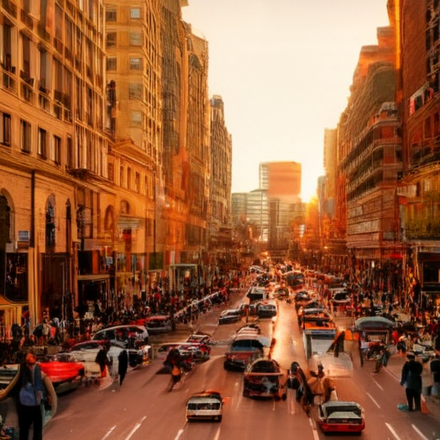} \\
    & \multicolumn{2}{c}{\small \emph{make it sunset}} \\
    \end{tabular}
    \end{minipage}
    \hfill
    \begin{minipage}[b]{\xxxwidth}
    \begin{tabular}{c  c c}
    \small Input & \small IP2P (200 steps) & \small \textbf{CoDi (Ours)} (1 step) \\
    \includegraphics[width=\xwidth]{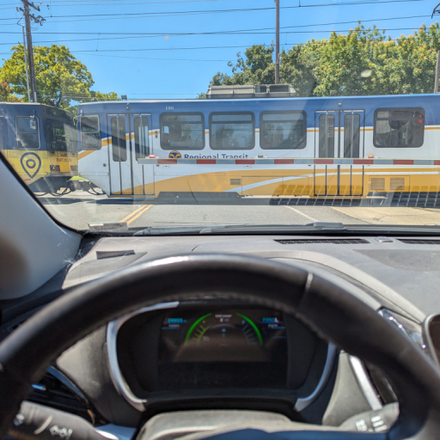} & \includegraphics[width=\xwidth]{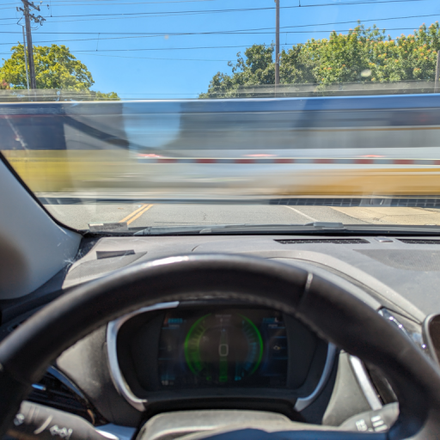} &
    \includegraphics[width=\xwidth]{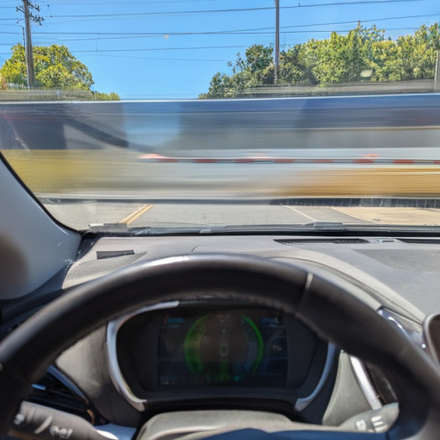} \\
    & \multicolumn{2}{c}{\small \emph{make it long exposure}} \\
    \end{tabular}
    \end{minipage}
    
    \begin{minipage}[b]{\xxxwidth}
    \begin{tabular}{c  c c}
    \small Input & \small IP2P (200 steps) & \small \textbf{CoDi (Ours)} (1 step) \\
    \includegraphics[width=\xwidth]{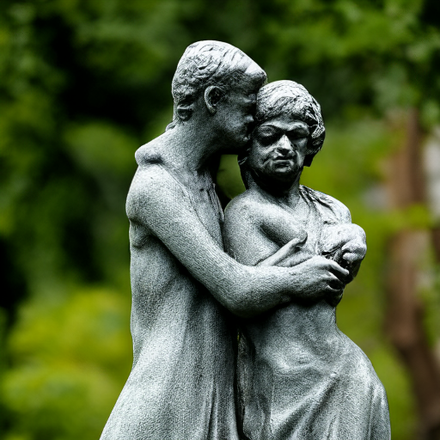} & \includegraphics[width=\xwidth]{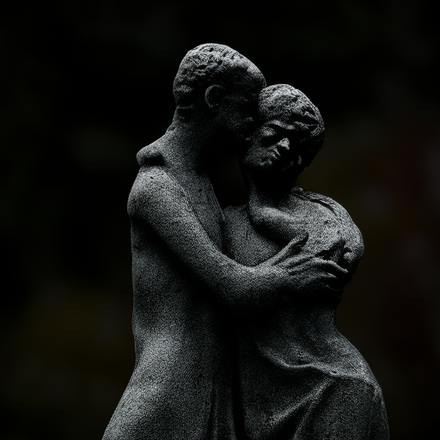} &
    \includegraphics[width=\xwidth]{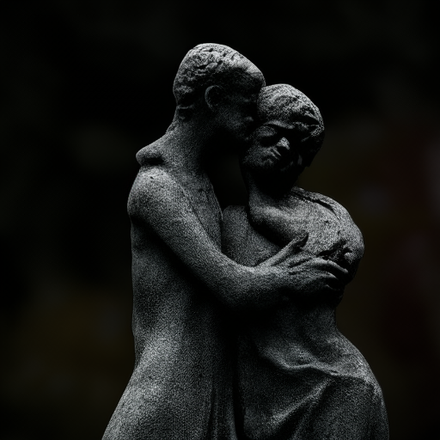} \\
    & \multicolumn{2}{c}{\small \emph{make it low key}} \\
    \end{tabular}
    \end{minipage}
    \hfill
    \begin{minipage}[b]{\xxxwidth}
    \begin{tabular}{c c c}
    \small Input & \small IP2P (200 steps) & \small \textbf{CoDi (Ours)} (1 step) \\
    \includegraphics[width=\xwidth]{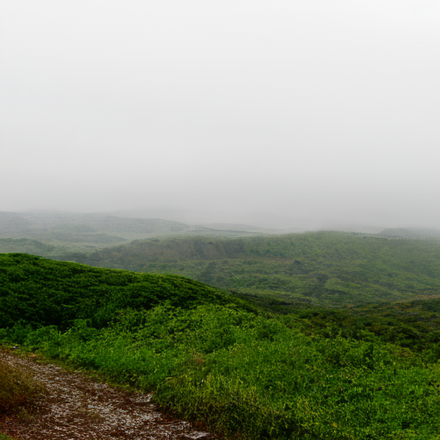} & \includegraphics[width=\xwidth]{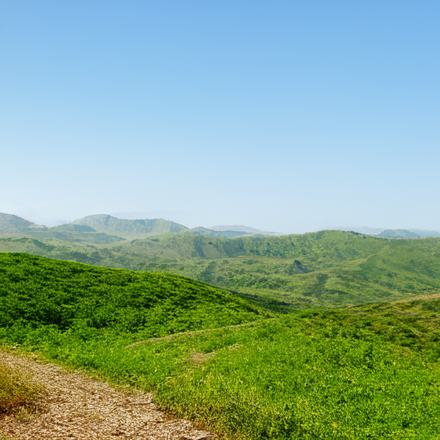} &
    \includegraphics[width=\xwidth]{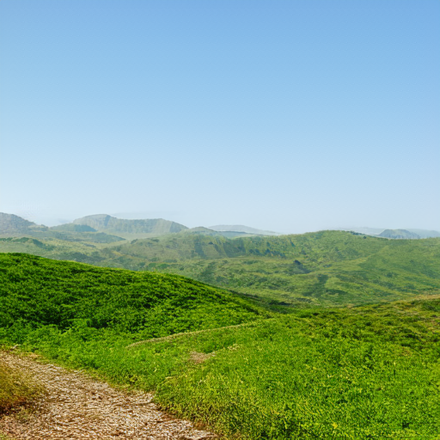} \\
    & \multicolumn{2}{c}{\small \emph{make it sunny}} \\
    \end{tabular}
    \end{minipage}

    \vspace{-.5\baselineskip}
    \caption{Generated edited image according to the input image and the instruction (bottom) from Instructed Pix2Pix (IP2P) sampled in 200 steps and ours sampled in 1 step. Please see our supplement for many more examples.}
    \label{fig:ip2p}
    \vspace{-1\baselineskip}
\end{figure*}

\section{Experiments}

We demonstrate the efficacy of our method on representative conditional generation tasks, including, real-world super-resolution~\citep{wang2022realesrgan}, depth-to-image generation~\citep{zhang2023adding}, and instructed image editing~\citep{brooks2023instructpix2pix}.
We utilize a pre-trained text-to-image latent diffusion models\footnote{We base our work on a version of Latent Diffusion Model trained on internal text-to-image data. It is comparable with StableDiffusion v1.4.} and conduct conditional distillation directly from the model.
Each of the compared methods, including the text-to-image pretraining, was independently trained for 8 days on 64 TPU-v4 pods.

\subsection{Results}

\noindent \textbf{Baselines.}
We compare our method with two previous SOTA diffusion distillation methods, \ie, consistency models (CM)~\cite{song2023consistency} and guided-distillation (GD)~\cite{meng2023distillation}.
We implement CM with ControlNet without freezing denoising U-Net, which leads to the same network architecture and the same number of parameters as ours.
For completeness, we consider two different ways of applying the tested distillation techniques, by first making the model conditional (fine-tuning first), or by first distilling the model and then making it conditional (distill first). A summary of the tested configurations is shown in~Table~\ref{tab:baselines}.
Additionally, we compare our method to recently introduced fast ODE solvers, including DPM-Solver~\cite{lu2022dpm} and DPM-Solver++~\cite{lu2022dpm2}.

\noindent \textbf{Real-world super-resolution.}
We evaluate our method on the challenging real-world super-resolution task, where the degradation is simulated using Real-ESRGAN pipeline~\citep{wang2021real}.
Following StablSR~\cite{wang2023exploiting}, we compare all methods on 3,000 randomly degraded image pairs.
The quantitative performance is shown in Table~\ref{tab:performance}.
The results demonstrate that our distilled method leads to a significant better performance than other distillation techniques.
Our method achieves better results than fine-tuned diffusion models that requires 50$\times$ more sampling setps.
Compared with the distilled model by applying the guided-distillation, our model outperforms it both quantitatively and qualitatively.
The visual comparison presented in Figure.~\ref{fig:srinpainting} also demonstrates the superiority of our method.

\noindent \textbf{Inpainting.}
Similar to the above super-resolution comparisons, we demonstrate our method on the inpainting task that conditioned on the masked image, as the quantitative performance shown in Table~\ref{tab:performance}.
Similar to Palette~\cite{saharia2022palette}, we apply random masks into ImageNet data~\cite{russakovsky2015imagenet} for both training and testing.
Note that we conduct experiments on the up-scaled images in a $512\times 512$ resolution, which is different than Palette in $256\times256$ resolution.
Even though we evaluate their results in the same resoltuion, their number can only be used for reference.

\noindent \textbf{Depth-to-image generation.}
\label{sec:d2i}
In order to demonstrate the generality of our method on less informative conditions, we apply our method in depth-to-image generation.
The task is usually conducted in parameter-efficient diffusion model finetuning~\citep{mou2023t2i, zhang2023adding}, which can demonstrate the capability of utilizing text-to-image generation priors.
As Figure~\ref{fig:depth} illustrated, our distilled model from the unconditional pretraining can effectively utilize the less informative conditions and generate matched images with more details.

\begin{figure*}[t]
\centering
\begin{minipage}[c]{.38\linewidth}
\captionsetup{type=table} %
    \centering
    \small
    \resizebox{\linewidth}{!}{
    \begin{tabular}{lccc}
        \toprule
         Methods & Params & FID & LPIPS \\
        \midrule
        LDMs & 865M & 29.266 & 0.3014  \\
         + ControlNet & 1.22B & 28.951 & 0.3049 \\ 
         \midrule
        PE-CoDi (Ours) & 364M & 25.214 & 0.2941 \\
        CoDi (Ours) & 1.22B & 19.637 & 0.2656 \\
        - distilling PF-ODE & 1.22B & 20.307 & 0.2733  \\
        - noise-consistency & 1.22B & 25.728 & 0.3252 \\
        \bottomrule
    \end{tabular}
    }
\caption{Impact of the network architecture and conditional distillation process, where all methods are using the same 4-step sampling.}\label{tab:ablation}
\end{minipage}
\begin{minipage}[c]{.60\linewidth}
\centering
\begin{subfigure}[b]{0.32\linewidth}
    \centering
    \includegraphics[height=1.12\linewidth]{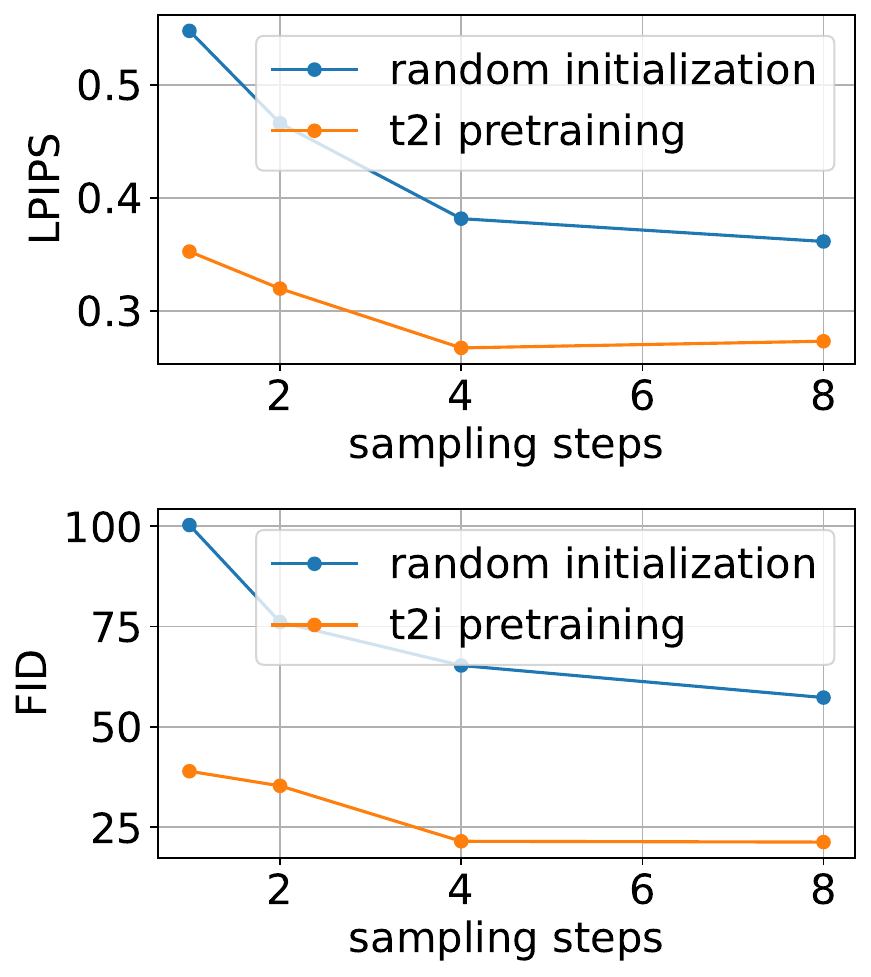}
\end{subfigure}
\begin{subfigure}[b]{0.32\linewidth}
    \centering
    \includegraphics[height=1.12\linewidth]{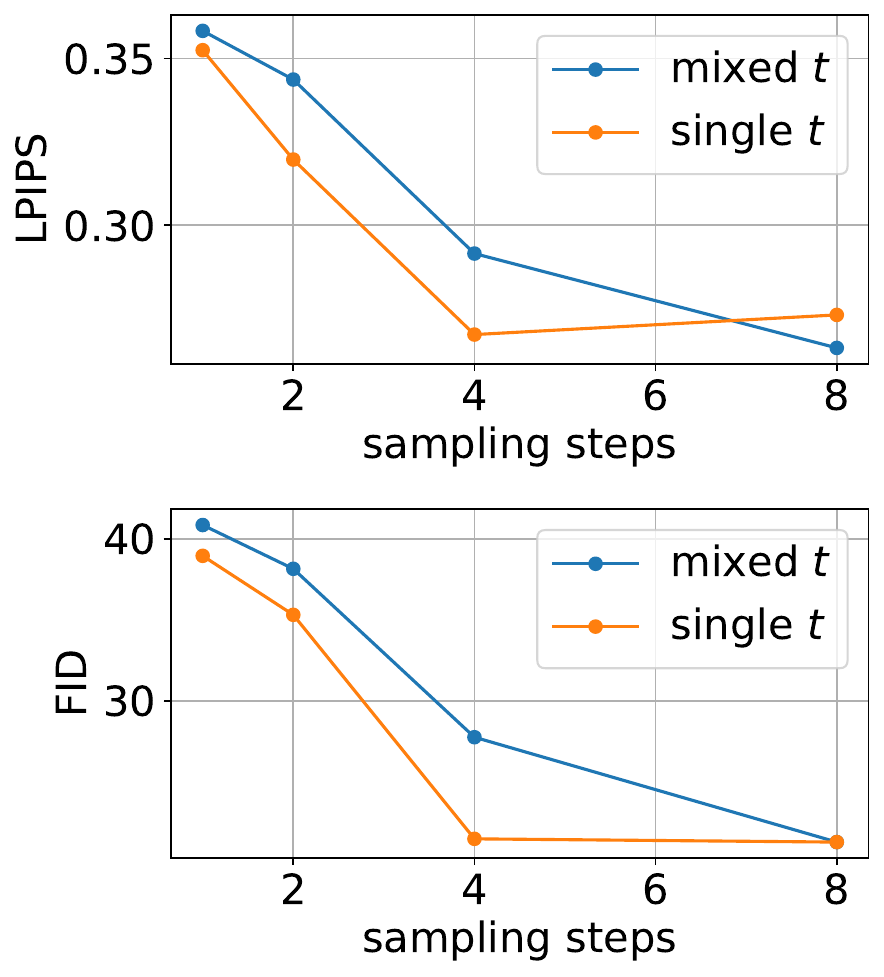}
\end{subfigure}
\begin{subfigure}[b]{0.32\linewidth}
    \centering
    \includegraphics[height=1.12\linewidth]{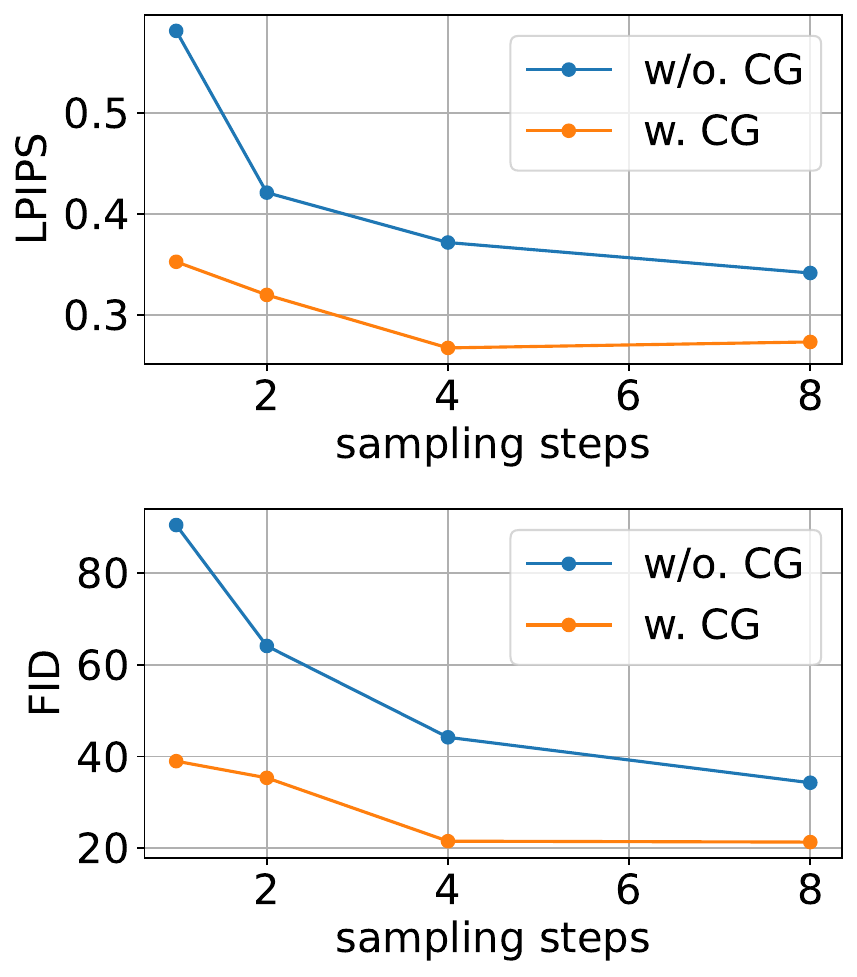}
\end{subfigure}
\vspace{-.5\baselineskip}
\caption{Ablations between alternative settings of our method.}
\label{fig:ablations}
\end{minipage}
\end{figure*}

\noindent \textbf{Instructed image editing.}
To demonstrate our conditional distillation capability on text-to-image generation, here we apply our method on text-instructed image editing data~\citep{brooks2023instructpix2pix} and compare our conditional distilled model with the InstructPix2Pix (IP2P) model.
As the results shown in Figure~\ref{fig:ip2p}, our single-step sampling result can achieve comparable visual quality to 200 steps of the IP2P model.
We experimentally find only small visual difference between the results from our single-step sampling and the 200 steps sampling.
We believe this suggests that the effect of the conditional guidance on distillation correlates with the similarity between the conditions and the target data, further demonstrating the effectiveness of our method.

\subsection{Ablations}
\label{sec:ablations}
Here we compare the performance of the aforementioned designs in our conditional distillation framework. Specifically we focus on the representative conditional generation task \emph{i.e.},  real-world super-resolution~\citep{wang2022realesrgan} that conditions on the low-resolution, noisy, blurry images.

\noindent \textbf{Network architecture and distillation process.}
To eliminate the impact of the architecture change, we compare our method with a baseline given by adding a ControlNet module trained on super-resolution without freezing the UNet. As Table~\ref{tab:ablation} shows, simply adopting a ControlNet module for super-resolution has negligible impact on the performance.
To evaluate the proposed conditional diffusion consistency, we removed the noise consistency term (\eqref{eq:loss}) and employed the training model in the PF-ODE instead of the frozen one as used in \cite{song2023consistency} formulation.
As shown in Table~\ref{tab:ablation}, adopting the distillation model PF-ODE and noise-space consistency have positive effects on the final results.
These comparisons demonstrate the superiority of our method without network architecture effects.

\noindent \textbf{Pretraining.}
To validate the effectiveness of leveraging pretraining in our model, we compare the results of random initialization with initialization from the pre-trained text-to-image model.
As shown in Figure~\ref{fig:ablations}, our method outperforms the random initialized counterpart by a large margin, thereby confirming that our strategy indeed utilizes the advantages of pretraining during distillation instead of simply learning from scratch.

\noindent \textbf{Sampling of~$\bz_t$.}
We empirically show that the way of sampling $\bz_t$ plays a crucial role in the distillation learning process.
Compared with the previous protocol~\citep{salimans2022progressive, meng2023distillation} that samples $\bz_t$ in different time $t$ in a single batch, we show that using a consistent time $t$ across different samples in a single batch leads to a better performance in our targeted 1-4 steps.
As the comparisons shown in Figure~\ref{fig:ablations}, the model trained with a single time $t$ (in a single batch) achieves better performance in both the visual quality (\emph{i.e.}, FID) and the accuracy (\emph{i.e.}, LPIPS) when the number of evaluations is increasing during inference.

\noindent \textbf{Conditional guidance.}
In order to demonstrate the importance of our proposed conditional guidance (CG) for distillation, which is claimed to be capable of regularizing the distillation process during training, we conduct comparisons between the setting of using the conditional guidance as $r=\| \bx - \hat{\bx}_\theta (\bz_t, c) \|^2_2$ and not using as $r=0$.
As the result shown in Figure~\ref{fig:ablations}, the conditional guidance improves both the fidelity of the generated results and visual quality.
We further observed that the distillation process will converge toward over-saturated direction without CG, which thus lower the FID metric. In contrast, our model avoids such a local minimum by using the proposed guidance loss.

\section{Conclusion}
We introduce a new framework for distilling an unconditional diffusion model into a conditional one that allows sampling with very few steps.
To the best of our knowledge, this is the first method that distills the conditional diffusion model from the unconditional pretraining in a single stage.
Compared with previous two-stage distillation and finetuning techniques, our method leads to better quality given the same number of (very few) sampling steps.
Our method also enables a new parameter-efficient distillation that allows different distilled models, trained for different tasks, to share most of their parameters. Only a few additional parameters are needed for each different conditional generation task. We believe the method can serve as a strong practical approach for accelerating large-scale conditional diffusion models.

\section{Acknowledgments.}
The authors would like to thank our colleagues Keren Ye and Chenyang Qi for reviewing the manuscript and providing valuable feedback.
We also extend our gratitude to Shlomi Fruchter, Kevin Murphy, Mohammad Babaeizadeh, and Han Zhang for their instrumental
contributions in facilitating the initial implementation of the latent diffusion models.

{
    \small
    \bibliographystyle{ieeenat_fullname}
    \bibliography{main_with_authors}
}

\input{appendix}

\end{document}

%% file: preamble.tex
\usepackage[dvipsnames]{xcolor}

%% file: math_commands.tex
\usepackage{amsmath,amsfonts,bm}

\def\eqref#1{equation~\ref{#1}}

\def\1{\bm{1}}

\def\vtheta{{\bm{\theta}}}

\DeclareMathAlphabet{\mathsfit}{\encodingdefault}{\sfdefault}{m}{sl}
\SetMathAlphabet{\mathsfit}{bold}{\encodingdefault}{\sfdefault}{bx}{n}

\newcommand{\bv}{\mathbf{v}}
\newcommand{\bw}{\mathbf{w}}
\newcommand{\bx}{\mathbf{x}}

\newcommand{\bz}{\mathbf{z}}

\newtheorem{theorem}{Theorem}

%% file: tables/regularization.tex
\begin{figure*}[ht]
    \centering
    \captionsetup[subfigure]{labelformat=empty}
    \raisebox{-0.2em}{
    \begin{subfigure}[b]{0.25\textwidth}
    \centering
    \includegraphics[width=\linewidth]{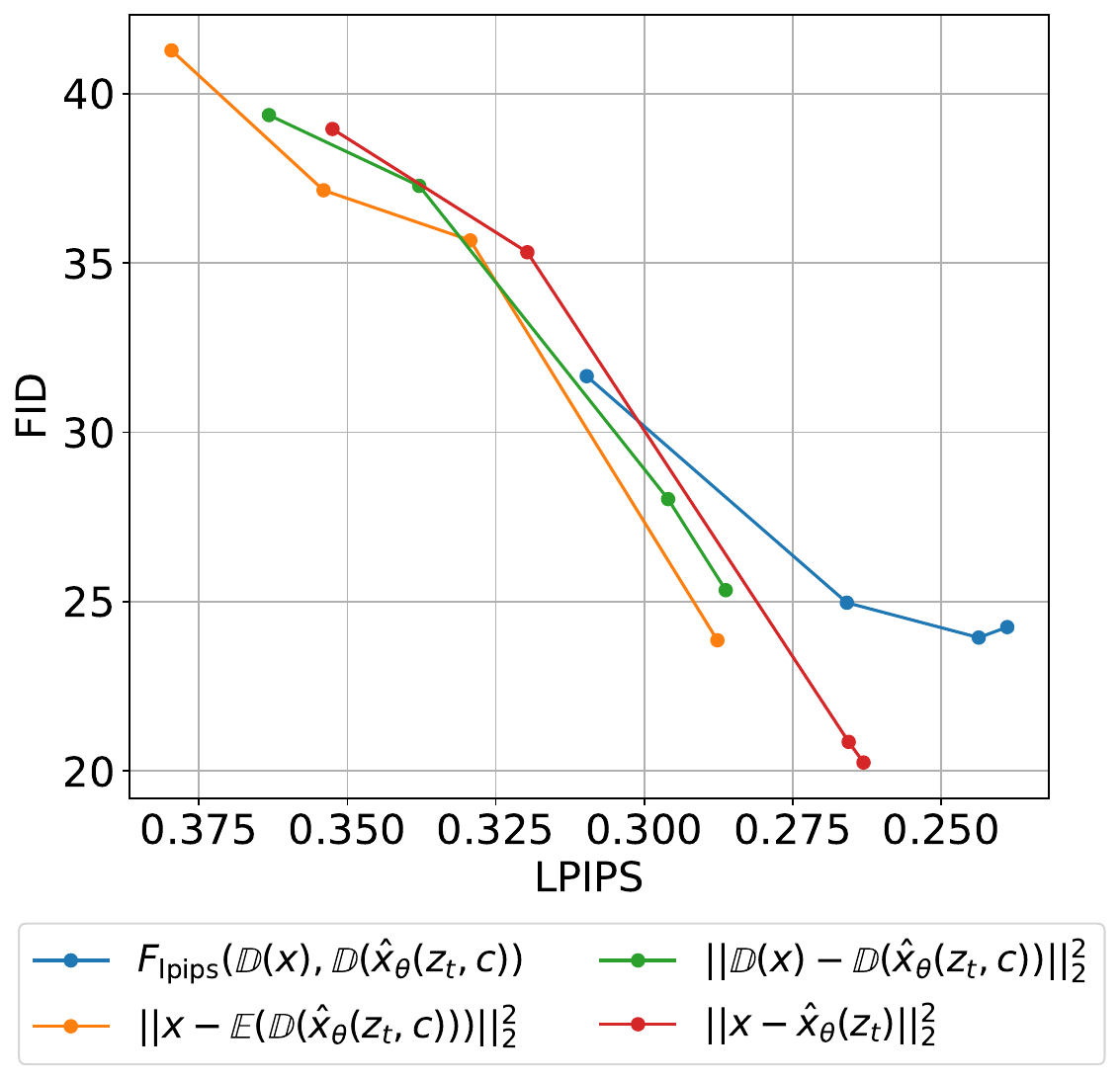}
    \end{subfigure}
    }
    \begin{subfigure}[b]{0.17\textwidth}
    \centering
    \includegraphics[width=\linewidth,cfbox=blue 1pt 0pt]{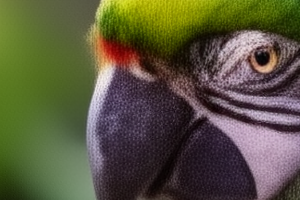}
    \includegraphics[width=\linewidth,cfbox=green 1pt 0pt]{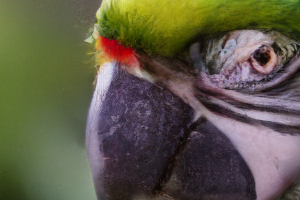}
    \end{subfigure}
    \begin{subfigure}[b]{0.17\textwidth}
    \centering
    \includegraphics[width=\linewidth,cfbox=blue 1pt 0pt]{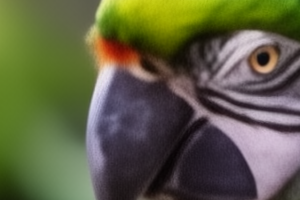}
    \includegraphics[width=\linewidth,cfbox=green 1pt 0pt]{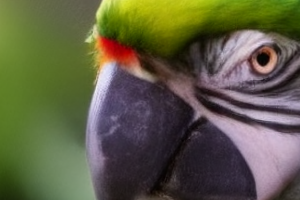}
    \end{subfigure}
    \begin{subfigure}[b]{0.17\textwidth}
    \centering
    \includegraphics[width=\linewidth,cfbox=blue 1pt 0pt]{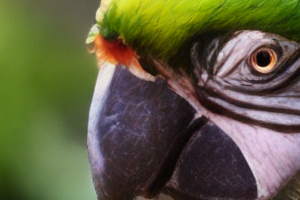}
    \includegraphics[width=\linewidth,cfbox=green 1pt 0pt]{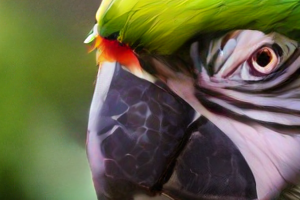}
    \end{subfigure}
    \begin{subfigure}[b]{0.17\textwidth}
    \centering
    \includegraphics[width=\linewidth,cfbox=blue 1pt 0pt]{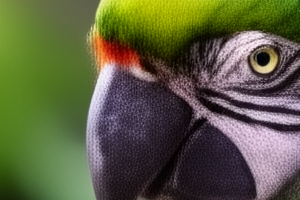}
    \includegraphics[width=\linewidth,cfbox=green 1pt 0pt]{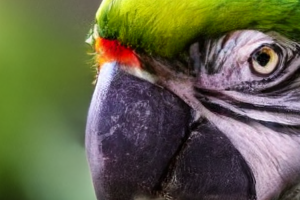}
    \end{subfigure}
    \hfill
    \caption{Sampled results between distilled models learned with alternative conditional guidance.
    Left curves shows the quantitative performance between the LPIPS and FID in $\{1,2,4,8\}$ steps.
    Right part show the visual results where each result comes from the 1 sampling step (top) or 4 sampling steps (bottom). The distance function from the left to right is $\| \bx - \mathbb{E}(\mathbb{D}(\hat{\bx}_\theta(\bz_t, c))) \|^2_2$, $\| \mathbb{D}(\bx) - \mathbb{D}(\hat{\bx}_\theta(\bz_t, c)) \|^2_2$,  $F_\mathrm{lpips}(\mathbb{D}(\bx), \mathbb{D}(\hat{\bx}_\theta(\bz_t, c))$, and our default $\| \bx - \hat{\bx}_\theta(\bz_t) \|^2_2$, respectively.}
    \label{fig:regularization}
    \vspace{-1\baselineskip}
\end{figure*}

%% file: appendix.tex
\onecolumn
\appendix
\newtheorem*{proposition-non}{Proposition}
\newtheorem*{remark-non}{Remark}

\newpage
\section{Discussion}
\noindent \textbf{Limitations.}
We have shown image conditions benefit our distillation learning. However, the distillation learning depends on the adapter architecture that introduces additional computation in our current framework.
As a future work, we would like to explore lightweight network architectures~\citep{li2023snapfusion} in our distillation technique to further reduce the inference latency.
Nevertheless, CoDI's significantly reduced sampling steps lead to lower latency. See the following table (measured in TPUv5) for a detailed comparison: 

\begin{table}[htbp]
    \centering
    \setlength{\tabcolsep}{8pt}
    \begin{tabular}{ccccc}
    Method & CoDi (4step) & ControlNet (4step) & LDMs (4step) & LDMs (50step)  \\
    \midrule
    Latency (ms) & 107 $\pm$ 3 & 107 $\pm$ 3 & 103 $\pm$ 2 & 977 $\pm$ 1 \\
    \bottomrule
    \end{tabular}
\end{table}

\noindent \textbf{Ethics statement.}
The diffusion distillation technique introduce in this work holds the promise of significantly enhancing the practicality of diffusion models in everyday applications such as consumer photography and artistic creation. While we are excited about the possibilities this model offers, we are also acutely aware of the possible risks and challenges associated with its deployment.
Our model's ability to generate realistic scenes could be misused for generating deceptive content. We encourage the research community and practitioners to prioritize privacy-preserving practices when using our method.

\section{Proofs}
\label{sec:proofs}
\subsection{Notations}
We use $\hat{\bv}_\theta(\cdot, \cdot)$ to denote a pre-trained diffusion model that learns the unconditional data distribution $\bx \sim p_\mathrm{data}$ with parameters $\theta$.
The signal prediction and the noise prediction transformed by \eqref{eq:v} are denoted by $\hat{\bx}_\theta(\cdot, \cdot)$ and $\hat{\epsilon}_\theta(\cdot, \cdot)$, and they share the same parameters $\theta$ with $\hat{\bv}_\theta(\cdot, \cdot)$.

\subsection{Self-consistency in Noise Prediction}
\begin{remark-non}
\label{prop:consistency}
If a diffusion model, parameterized by $\hat{\bv}_\theta(\bz_t, t)$, satisfies the self-consistency property on the noise prediction $\hat{\epsilon}_\theta(\bz_t, t) = \alpha_t \hat{\bv}_\theta(\bz_t, t) + \sigma_t \bz_t$, then it also satisfies the self-consistency property on the signal prediction  $\hat{\bx}_\theta(\bz_t, t) = \alpha_t \bz_t - \sigma_t \hat{\bv}_\theta(\bz_t, t)$.
\end{remark-non}

\begin{proof}
The diffusion model that satisfies the self-consistency in the noise prediction implies:
\begin{equation}
    \begin{aligned}
        \hat{\epsilon}_\theta(\bz_{t'}, t') &= \hat{\epsilon}_\theta(\bz_t, t),  \\
        \alpha_{t'} \hat{\bv}_\theta(\bz_{t'}, t') + \sigma_{t'} \bz_{t'} &= \alpha_t \hat{\bv}_\theta(\bz_t, t) + \sigma_t \bz_t, \\
        \hat{\bv}_\theta(\bz_{t'}, t') &= \frac{\alpha_t \hat{\bv}_\theta(\bz_t, t) + \sigma_t \bz_t - \sigma_{t'} \bz_{t'}}{\alpha_{t'}},
    \end{aligned}
    \label{eq:vequal}
\end{equation}

Based on the above equivalence, the transformation between the signal prediction $\bx_\theta (\bz_{t'}, t')$ and $\bx_\theta (\bz_{t}, t)$ by using the update ruler in \eqref{eq:ddim} and the reparameterization trick is: 
\begin{align*}
\bx_\theta (\bz_{t'}, t') &= \alpha_{t'} \bz_{t'} - \sigma_{t'} \hat{\bv}_\theta (\bz_{t'}, t')\\
&= \alpha_{t'} \bz_{t'} - \sigma_{t'} \frac{\alpha_t \hat{\bv}_\theta(\bz_t, t) + \sigma_t \bz_t - \sigma_{t'} \bz_{t'}}{\alpha_{t'}}  && \text{ // integrating \eqref{eq:vequal} }\\
&=\frac{ \alpha_{t'}^2\bz_{t'} - \sigma_{t'} \alpha_t \hat{\bv}_\theta (\bz_t, t) - \sigma_{t'}\sigma_t \bz_t + \sigma_{t'}^2\bz_{t'}}{\alpha_{t'}} \\
&=\frac{ (1 - \sigma_{t'}^2)\bz_{t'} - \sigma_{t'} \alpha_t \hat{\bv}_\theta (\bz_t, t) - \sigma_{t'}\sigma_t \bz_t + \sigma_{t'}^2\bz_{t'}}{\alpha_{t'}} \\
&=\frac{ \bz_{t'} - \sigma_{t'}(\alpha_t \hat{\bv}_\theta (\bz_t, t) + \sigma_t \bz_t)}{\alpha_{t'}}\\
&=\frac{ \bz_{t'} - \sigma_{t'}(\hat{\epsilon}_\theta(\bz_t, t))}{\alpha_{t'}} && \text{ // transformed with \eqref{eq:v}}\\
&=\frac{\alpha_{t'}\bx_\theta(\bz_t, t) + \sigma_{t'}\hat{\epsilon}_\theta(\bz_t, t) - \sigma_{t'}(\hat{\epsilon}_\theta(\bz_t, t))}{\alpha_{t'}} && \text{ // update ruler \eqref{eq:vddim} of DDIM} \\
&= \bx_\theta(\bz_t, t). 
\end{align*}
The derived equivalence shows that enforcing the self-consistency in the noise prediction, which is implemented by learning to minimize our distillation loss in \eqref{eq:loss}, enforces the self-consistency in the signal prediction and can distill the pre-trained diffusion model.
\end{proof}

\section{Difference between Consisntecy Models}
\begin{algorithm}[H]
\caption{Conditional Diffusion Distillation (CDD)}
\label{alg:distillation}
\begin{algorithmic}
\STATE {\textbf{Input:}} conditional data $(\bx, c) \sim p_\mathrm{data}$, adapted diffusion model $\hat{\bw}_\theta(\bz_t, c, t)$, learning rate $\eta$, distance functions $d_\epsilon(\cdot, \cdot)$ and $d_\bx(\cdot, \cdot)$, and EMA $\gamma$ 
    \STATE $\vtheta^- \gets \vtheta$
    \COMMENT{target network initlization}
    \REPEAT
    \STATE Sample $(\bx, c) \sim p_\mathrm{data}$ and $t \sim [\Delta t, T]$
    \COMMENT{empirically $\Delta t = 1$}
    \STATE Sample $\epsilon \sim \mathcal{N}(0, \mathbf{I})$    
    \State $s \gets t - \Delta t$ 
    \STATE Sample $\bz_t \gets \alpha_t \bx + \sigma_t \epsilon$
    \STATE {\setlength{\fboxsep}{0pt}\colorbox{pink}{~-~~$\hat{\bx}_t \gets \alpha_t \bz_t  - \sigma_t \Phi(\bz_t, c, t)$}}   
    \STATE {\setlength{\fboxsep}{0pt}\colorbox{pink}{~-~~$\hat{\epsilon}_t \gets \alpha_t \Phi(\bz_t, c, t) + \sigma_t \bz_t$}}
    
    \STATE {\setlength{\fboxsep}{0pt}\colorbox{lime}{~+~~$\hat{\bx}_t \gets \alpha_t \bz_t  - \sigma_t \hat{\bw}_\theta(\bz_t, c, t)$}}
    \COMMENT{signal prediction in \eqref{eq:v}}
    \STATE {\setlength{\fboxsep}{0pt}\colorbox{lime}{~+~~$\hat{\epsilon}_t \gets \alpha_t \hat{\bw}_\theta(\bz_t, c, t) + \sigma_t \bz_t$}}
    \COMMENT{noise prediction in \eqref{eq:v}}
     \STATE $\hat{\bz}_s \gets \alpha_s \hat{\bx}_t + \sigma_s \hat{\epsilon}_t$ 
    \COMMENT{update rule in \eqref{eq:vddim}}
    \STATE {\setlength{\fboxsep}{0pt}\colorbox{pink}{~-~~$\hat{x}'_t \gets \alpha_t \bw_{\theta}(\bz_t, c, t) + \sigma_t \bz_t$}}  
    \STATE {\setlength{\fboxsep}{0pt}\colorbox{pink}{~-~~$\hat{x}'_s \gets \alpha_t \bw_{\theta^-}(\hat{\bz}_s, c, s) + \sigma_s \hat{\bz}_s$}}
    
    \STATE {\setlength{\fboxsep}{0pt}\colorbox{lime}{~+~~$\hat{\epsilon}_s \gets \alpha_s \bw_{\theta^-}(\hat{\bz}_s, c, t) + \sigma_s \hat{\bz}_s$}}  
    \COMMENT{noise prediction in \eqref{eq:v}}
    \STATE {\setlength{\fboxsep}{0pt}\colorbox{pink}{~-~~$
        \mathcal{L}(\theta, \theta^{-}) \gets  d_\bx(\hat{x}'_t, \hat{x}'_s)
    $}}
    \STATE {\setlength{\fboxsep}{0pt}\colorbox{lime}{~+~~$
        \mathcal{L}(\theta, \theta^{-}) \gets d_\epsilon(\hat{\epsilon}_t, \hat{\epsilon}_s) + d_\bx(\bx, \hat{\bx}_t)
    $}}
    \STATE $\vtheta \gets \vtheta - \eta \nabla_\vtheta \mathcal{L}(\vtheta, \vtheta^{-})$  
    \STATE $\vtheta^- \gets \operatorname{stopgrad}(\gamma \vtheta^- + (1-\gamma) \vtheta$)
    \COMMENT{exponential moving average}
    \UNTIL{convergence}
 \end{algorithmic}
 \end{algorithm}

\section{Implementation Details}
\paragraph{Skip Connections.}
We implement the skip connections as follows, which is same as the consistency models~\cite{song2023consistency} and EDMs~\cite{karras2022elucidating} for satisfying the boundary condition but $f_\phi$ could be either the signal prediction or noise prediction:
\begin{equation}
    f'_\phi (\bz_t, t) = c_\mathrm{skip}(t) \bx + c_\mathrm{out}(t) f_\phi (\bz_t, t),
    \label{eq:boundary}
\end{equation}
where
\begin{equation}
    c_\mathrm{skip}(t) = \frac{\sigma_\mathrm{data}}{t^2 + \sigma^2_\mathrm{data}}, c_\mathrm{out}(t) = \frac{\sigma_\mathrm{data} t}{\sqrt{t^2 + \sigma^2_\mathrm{data}}}.
\end{equation}
We use $\sigma_\mathrm{data}=0.5$.

\clearpage
\section{Additional results}

\begin{figure*}[ht]
    \centering
    \def\xwidth{0.12\linewidth}
    \def\ywidth{0.12\linewidth}
    \setlength{\tabcolsep}{1pt}
    \renewcommand\arraystretch{1.1}
    \begin{tabular}[t]{c c c c c c c c}\\
    \includegraphics[width=\xwidth]{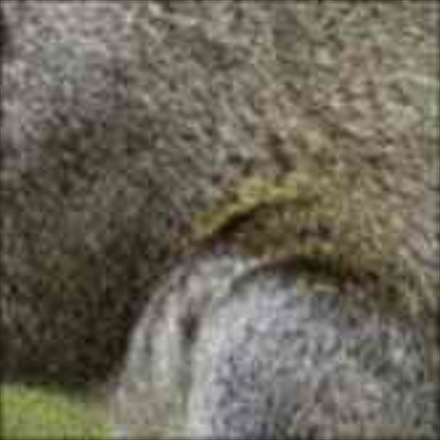} 
    & \includegraphics[width=\xwidth]{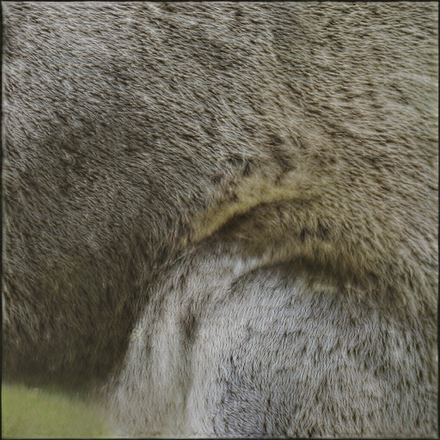} 
    & \includegraphics[width=\xwidth]{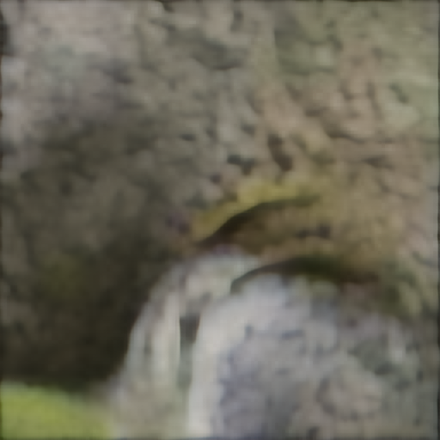} 
    & \includegraphics[width=\xwidth]{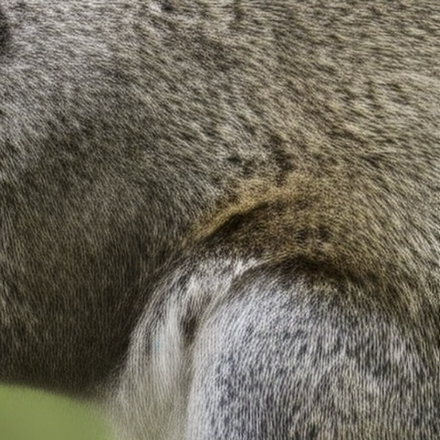} 
    & \includegraphics[width=\xwidth]{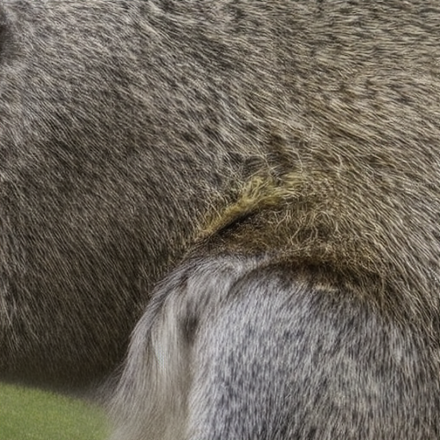} 
    & \includegraphics[width=\xwidth]{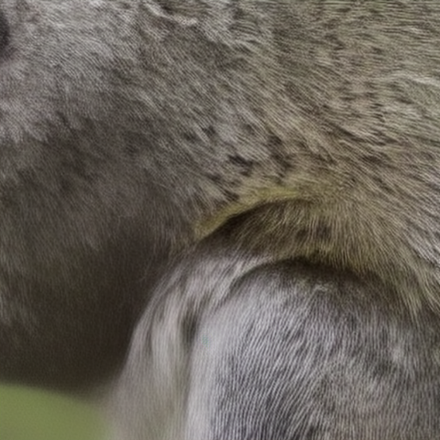}
    & \includegraphics[width=\xwidth]{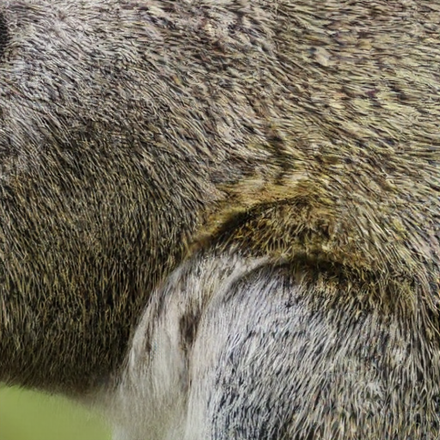}
    & \includegraphics[width=\xwidth]{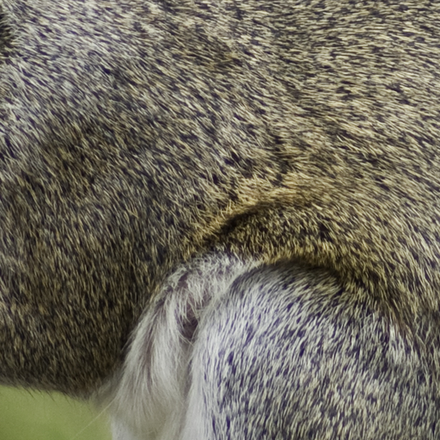}
    \\
    \includegraphics[width=\xwidth]{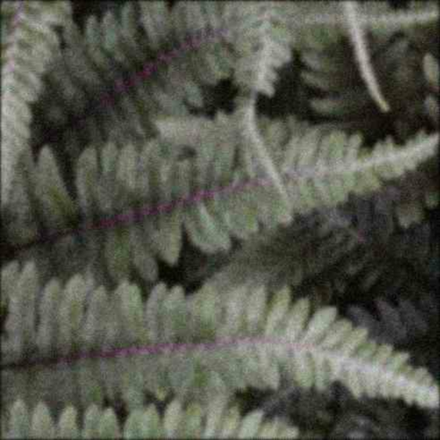} 
    & \includegraphics[width=\xwidth]{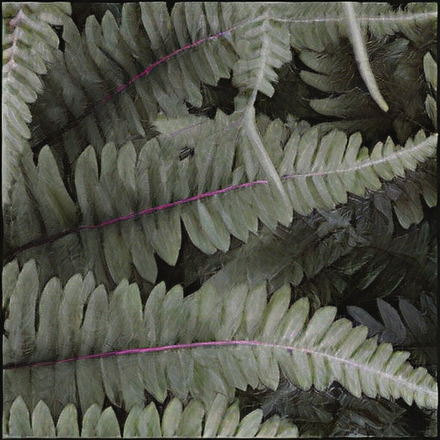} 
    & \includegraphics[width=\xwidth]{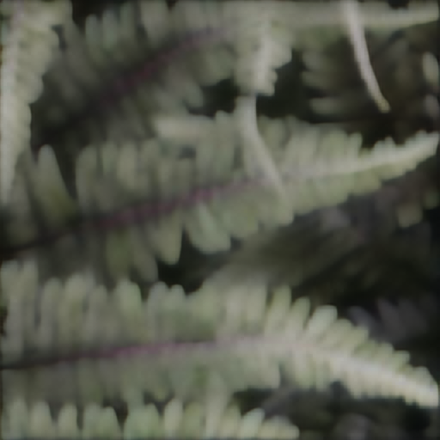} 
    & \includegraphics[width=\xwidth]{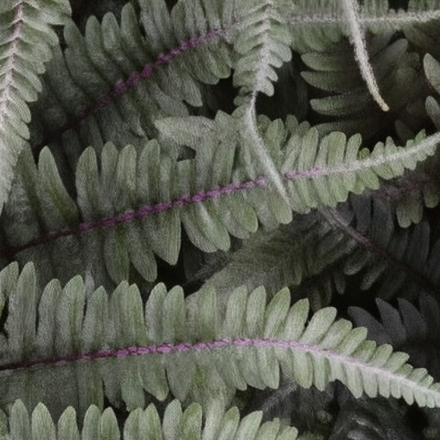} 
    & \includegraphics[width=\xwidth]{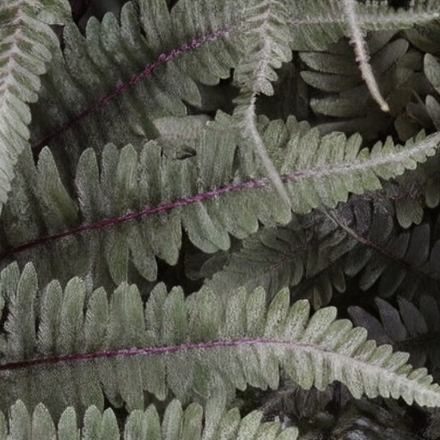} 
    & \includegraphics[width=\xwidth]{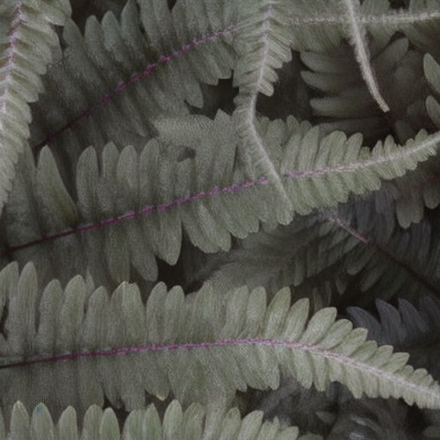}
    & \includegraphics[width=\xwidth]{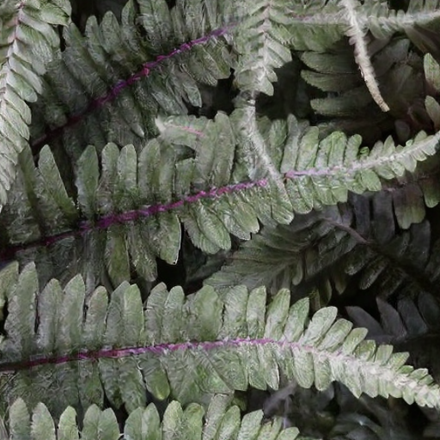}
    & \includegraphics[width=\xwidth]{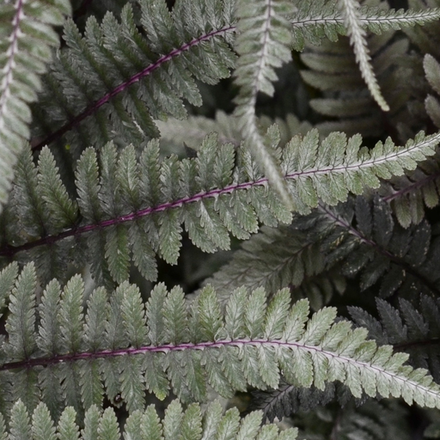}
    \\
    \includegraphics[width=\xwidth]{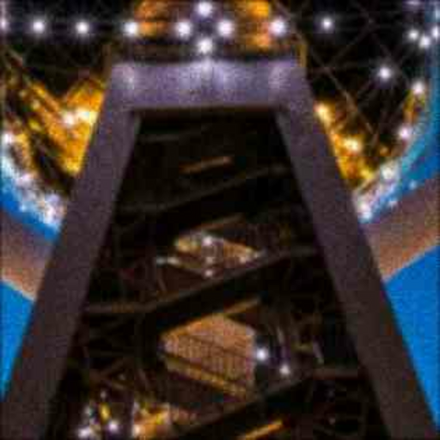} 
    & \includegraphics[width=\xwidth]{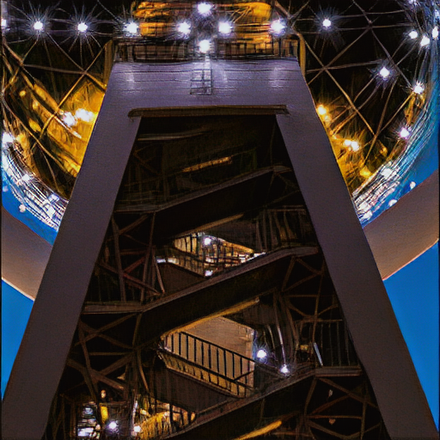}
    & \includegraphics[width=\xwidth]{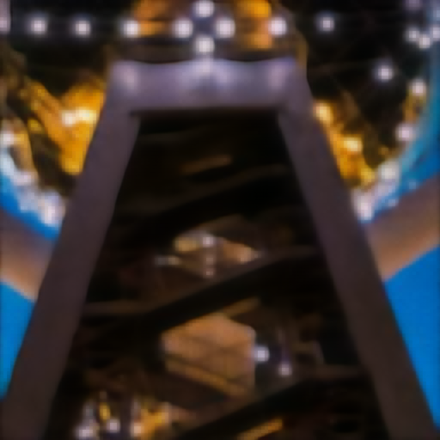} 
    & \includegraphics[width=\xwidth]{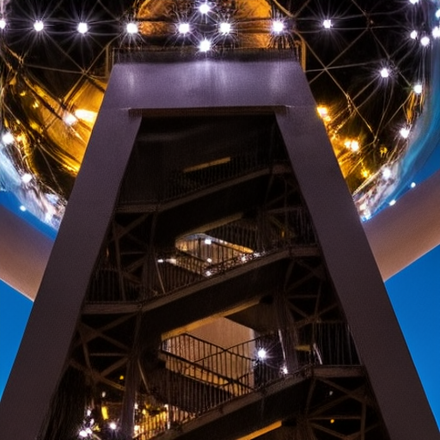} 
    & \includegraphics[width=\xwidth]{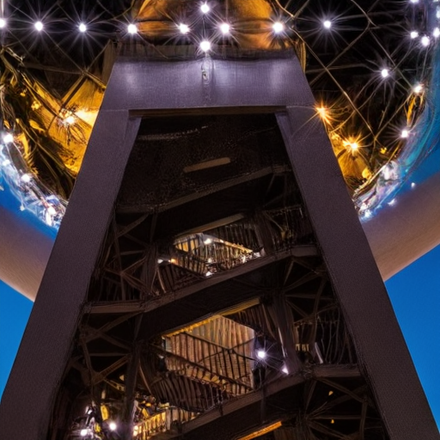} 
    & \includegraphics[width=\xwidth]{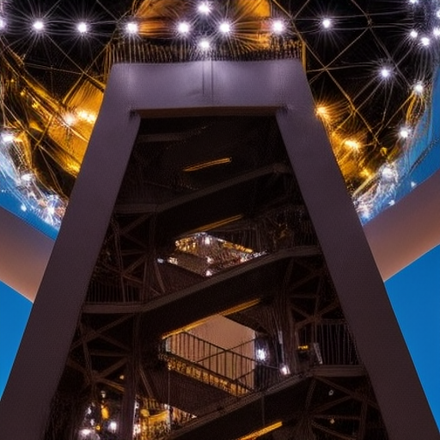}
    & \includegraphics[width=\xwidth]{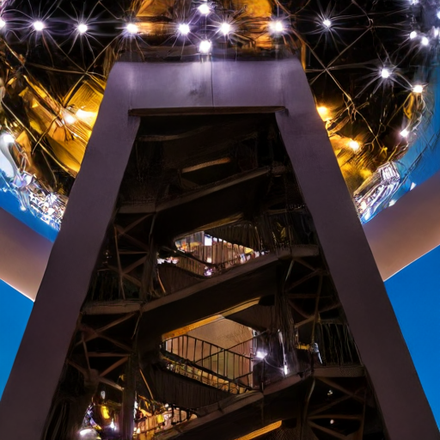}
    & \includegraphics[width=\xwidth]{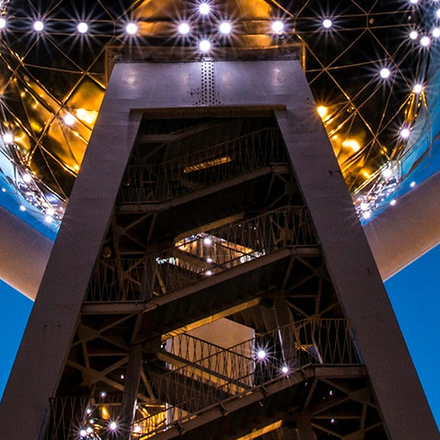}
    \\
    \includegraphics[width=\xwidth]{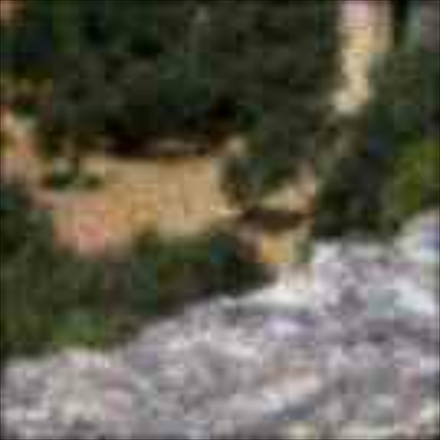} 
    & \includegraphics[width=\xwidth]{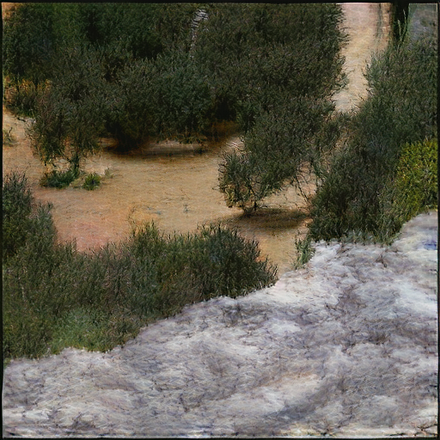}
    & \includegraphics[width=\xwidth]{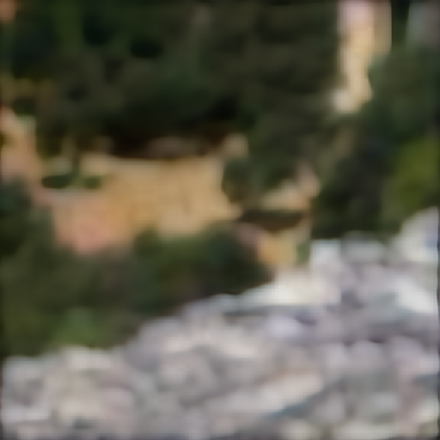} 
    & \includegraphics[width=\xwidth]{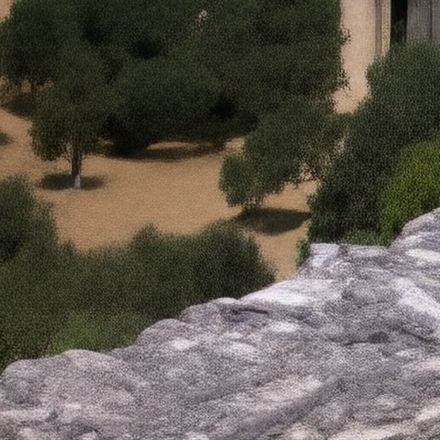} 
    & \includegraphics[width=\xwidth]{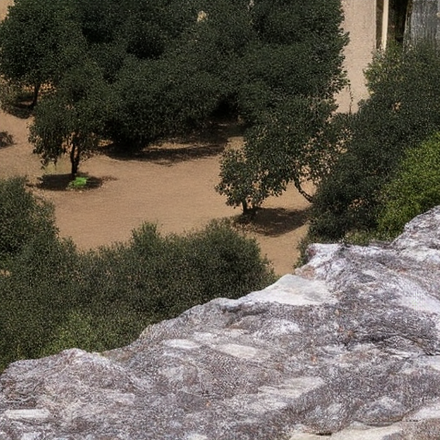} 
    & \includegraphics[width=\xwidth]{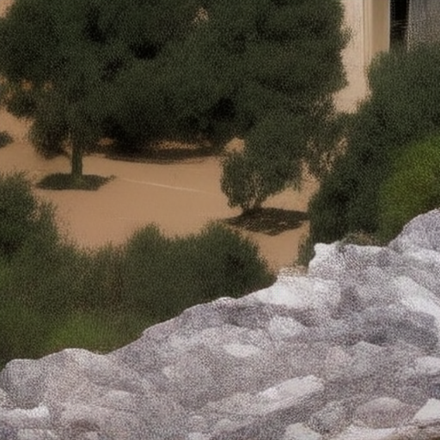}
    & \includegraphics[width=\xwidth]{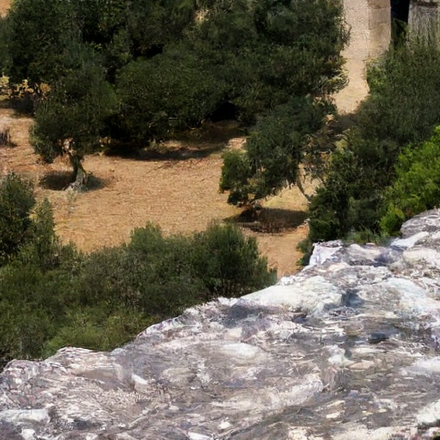}
    & \includegraphics[width=\xwidth]{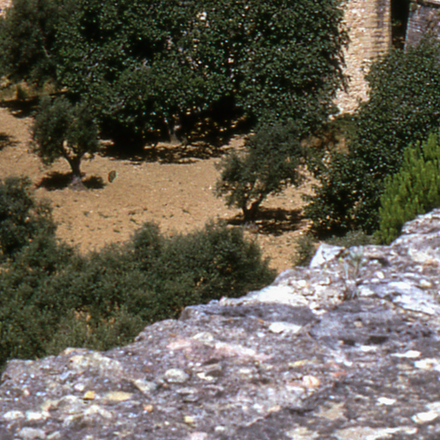}
    \\
    \includegraphics[width=\xwidth]{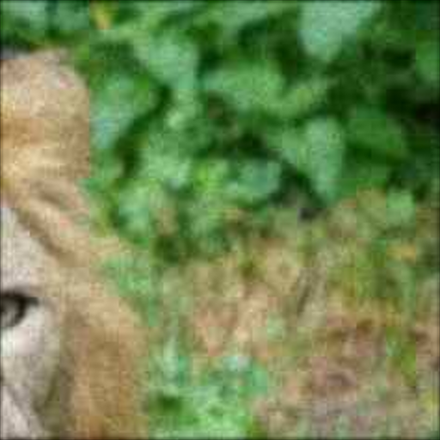} 
    & \includegraphics[width=\xwidth]{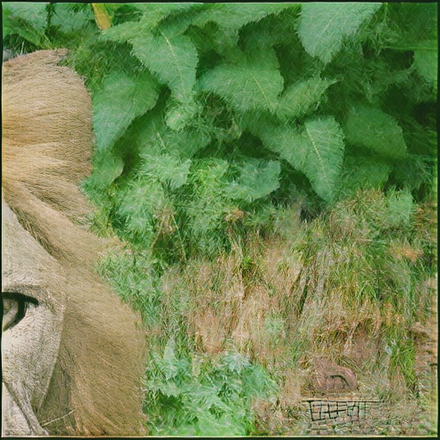}
    & \includegraphics[width=\xwidth]{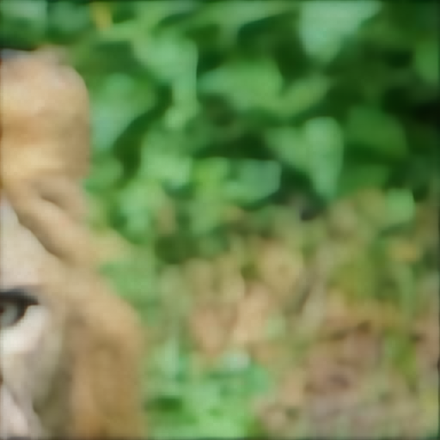} 
    & \includegraphics[width=\xwidth]{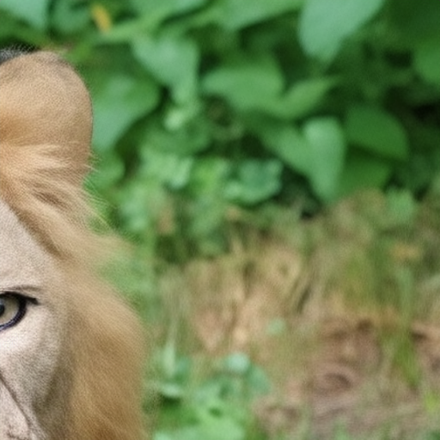} 
    & \includegraphics[width=\xwidth]{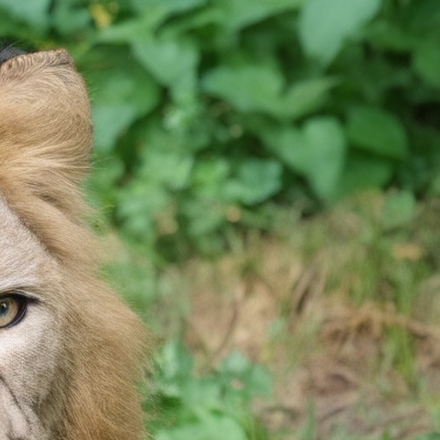} 
    & \includegraphics[width=\xwidth]{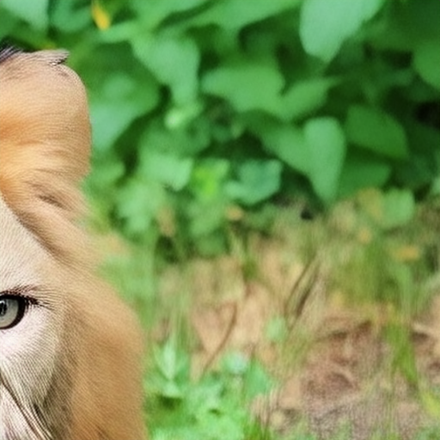}
    & \includegraphics[width=\xwidth]{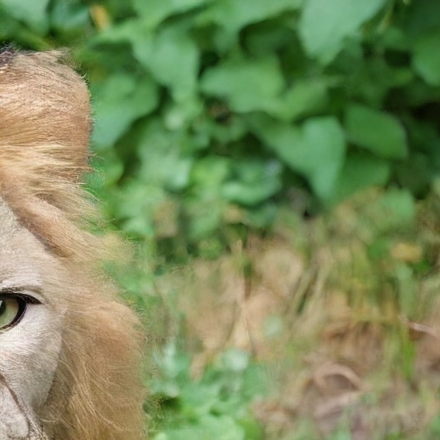}
    & \includegraphics[width=\xwidth]{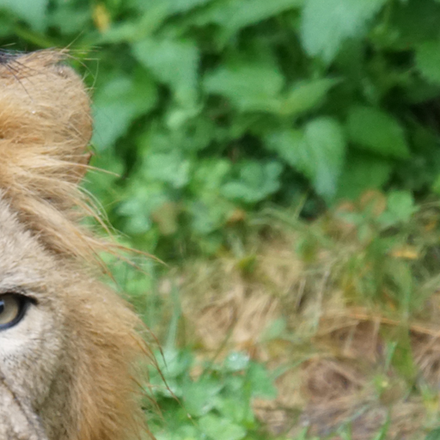}
    \\
    \includegraphics[width=\xwidth]{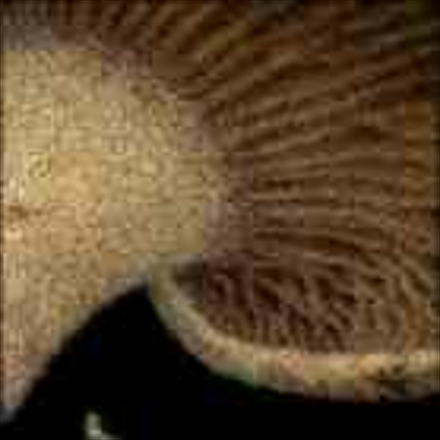} 
    & \includegraphics[width=\xwidth]{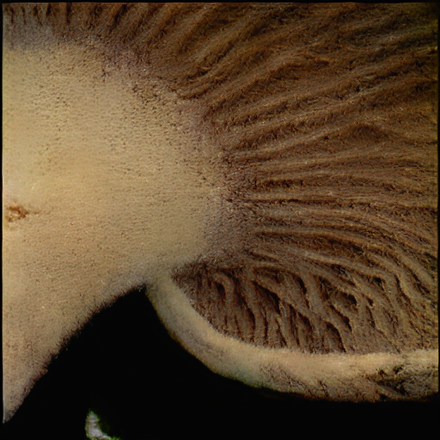}
    & \includegraphics[width=\xwidth]{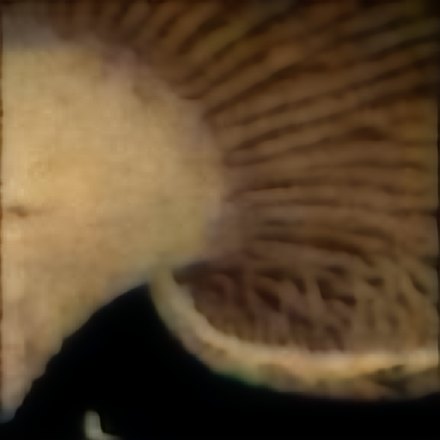} 
    & \includegraphics[width=\xwidth]{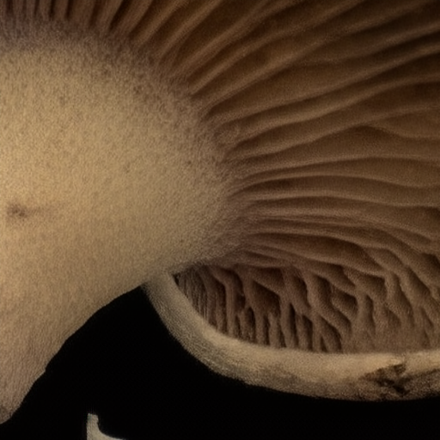} 
    & \includegraphics[width=\xwidth]{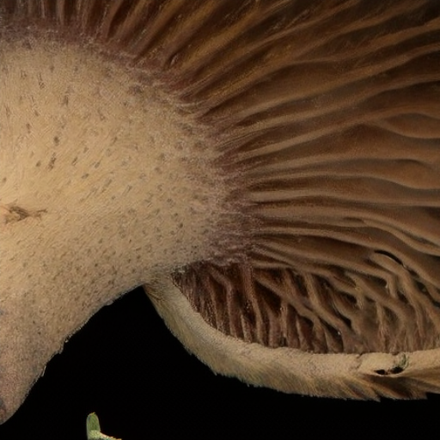} 
    & \includegraphics[width=\xwidth]{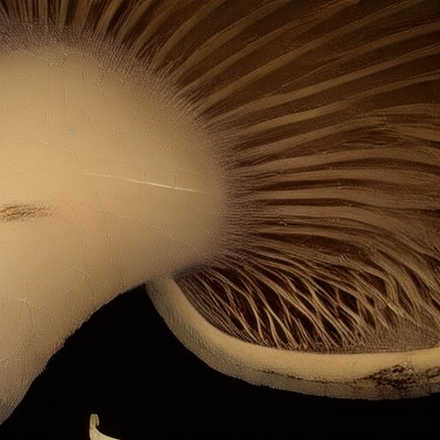}
    & \includegraphics[width=\xwidth]{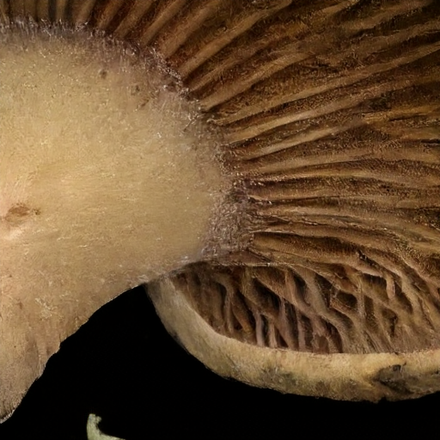}
    & \includegraphics[width=\xwidth]{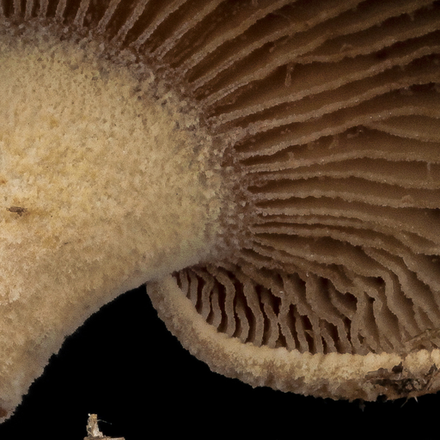}
    \\
    \includegraphics[width=\xwidth]{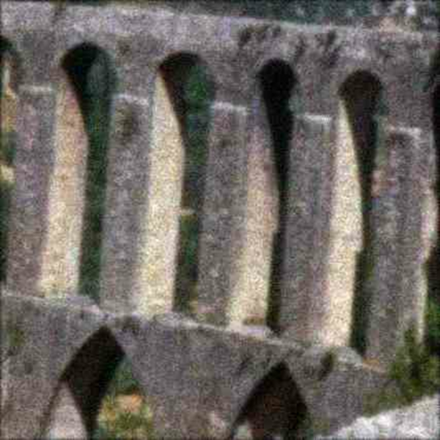} 
    & \includegraphics[width=\xwidth]{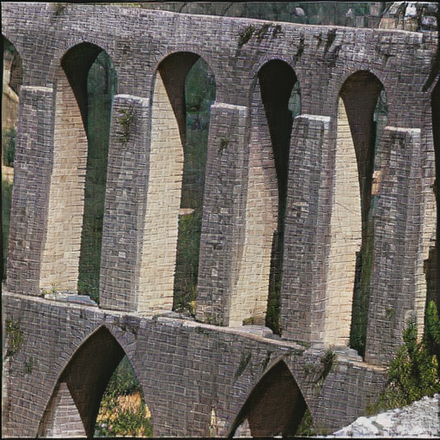}
    & \includegraphics[width=\xwidth]{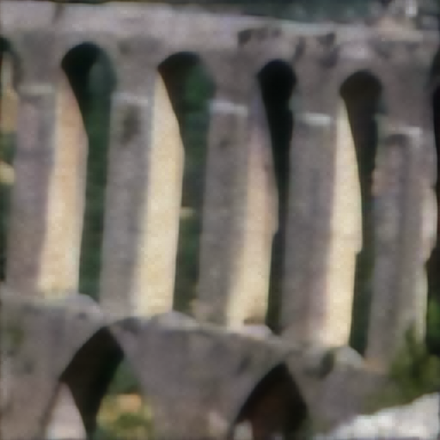} 
    & \includegraphics[width=\xwidth]{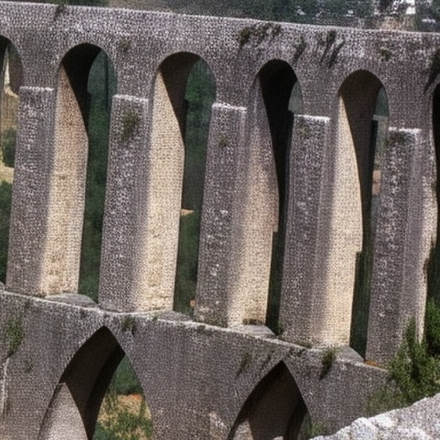} 
    & \includegraphics[width=\xwidth]{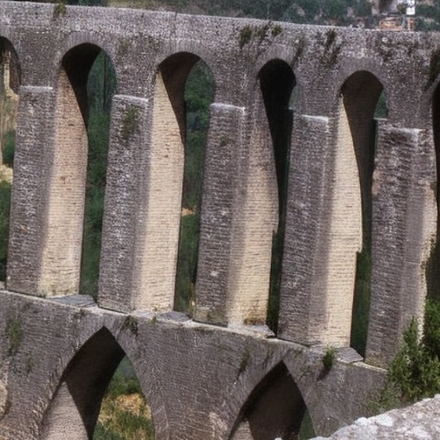} 
    & \includegraphics[width=\xwidth]{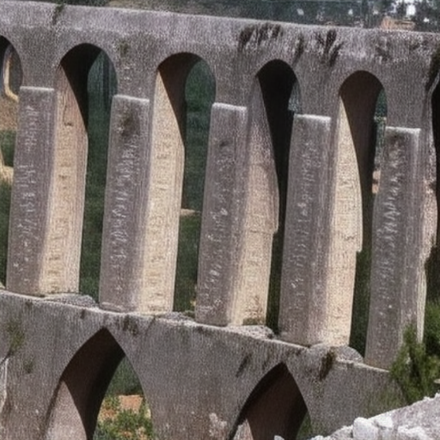}
    & \includegraphics[width=\xwidth]{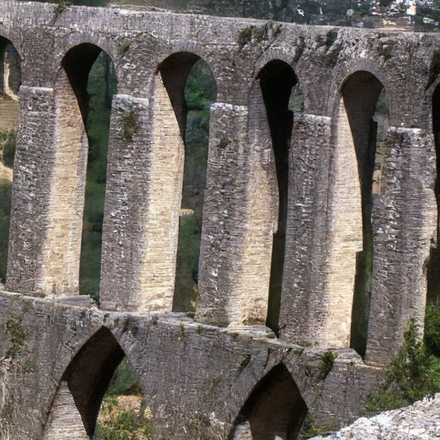}
    & \includegraphics[width=\xwidth]{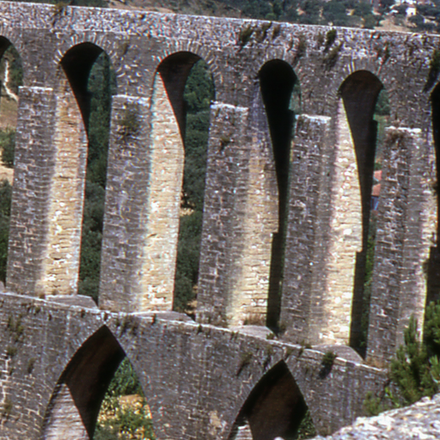}
    \\
    \includegraphics[width=\xwidth]{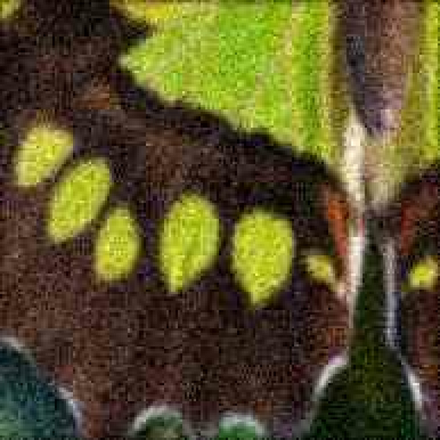} 
    & \includegraphics[width=\xwidth]{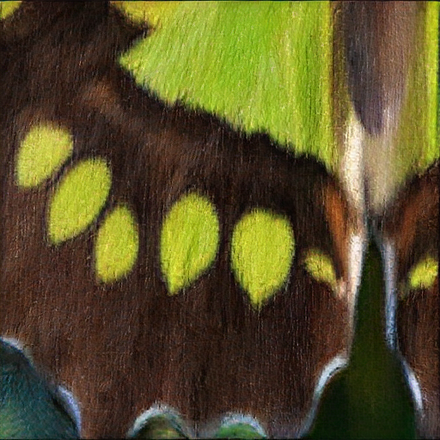}
    & \includegraphics[width=\xwidth]{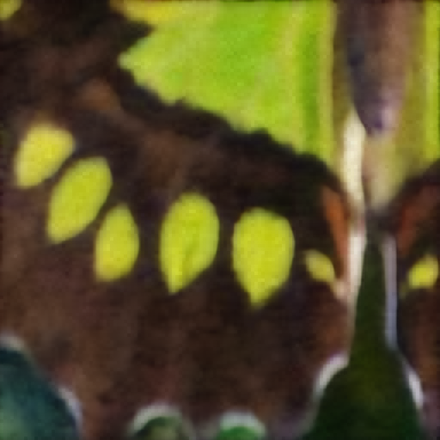} 
    & \includegraphics[width=\xwidth]{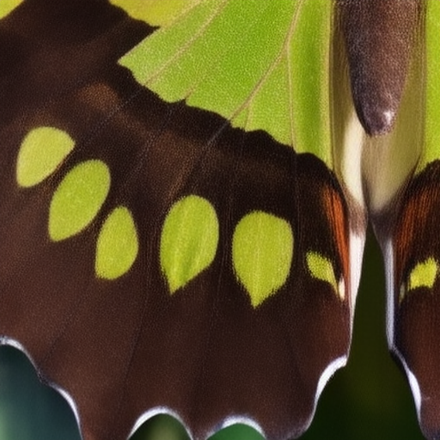} 
    & \includegraphics[width=\xwidth]{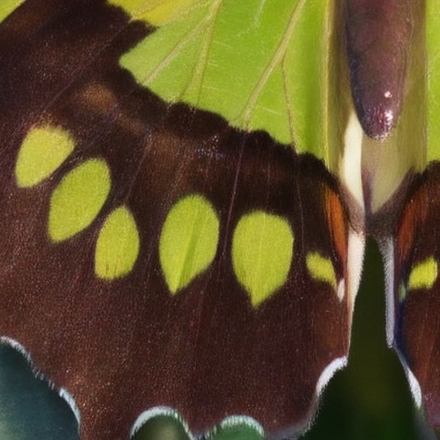} 
    & \includegraphics[width=\xwidth]{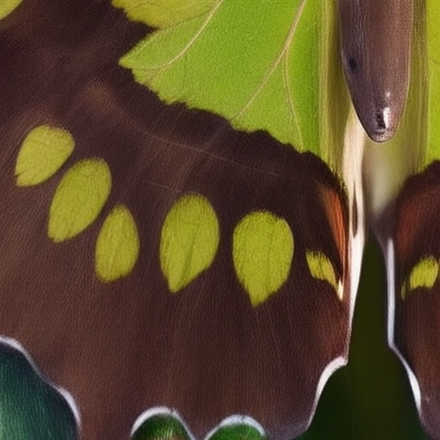}
    & \includegraphics[width=\xwidth]{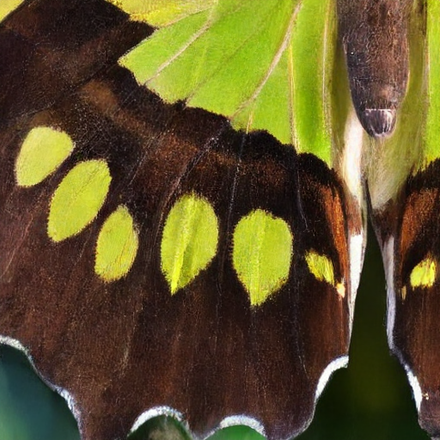}
    & \includegraphics[width=\xwidth]{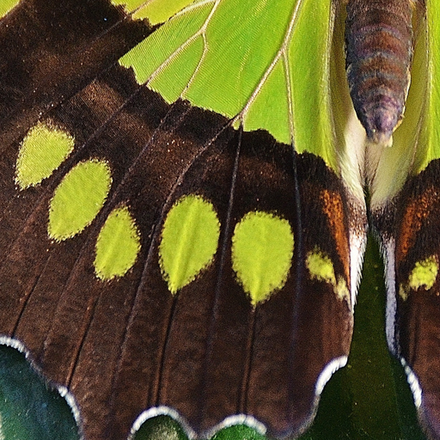}
    \\
    \scriptsize LR & \scriptsize StableSR & \scriptsize DiffIR  & \scriptsize LDMs (4 steps)  & \scriptsize GD-II (4 steps) & \scriptsize CM-II (4 steps) & \scriptsize CoDi (Ours) & \scriptsize HR \scriptsize  
    \end{tabular}
    \caption{Visual comparisons of various diffusion-based methods on the simulated real-world super-resolution benchmark. The input of all methods is a `Bicubic'-upsampled image.}
\end{figure*}

\begin{figure}[htbp]
    \centering
    \setlength{\tabcolsep}{4pt}
    \begin{tabular}{c | c c c}
    \small \textbf{Input} & \small \textbf{IP2P} (200 steps) & \small \textbf{Ours} (1 step) & \small \textbf{Ours} (4 step) \\
    \includegraphics[width=.19\linewidth]{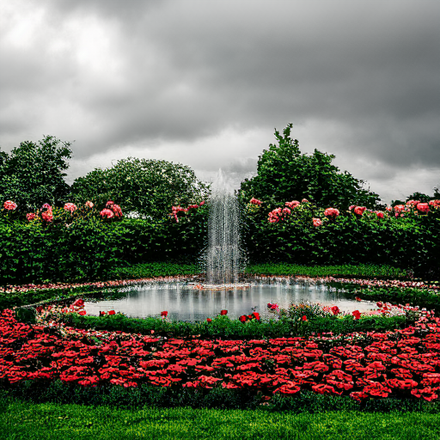} & \includegraphics[width=.19\linewidth]{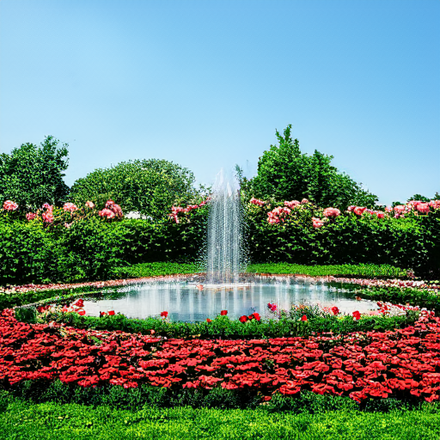} &
    \includegraphics[width=.19\linewidth]{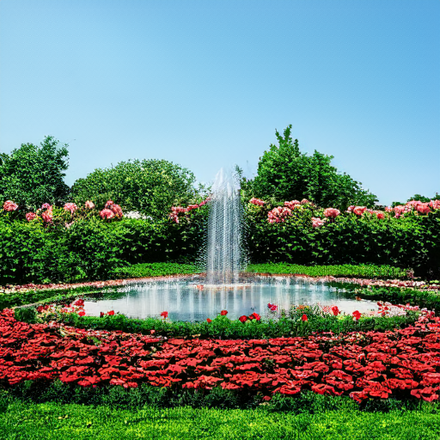} &    \includegraphics[width=.19\linewidth]{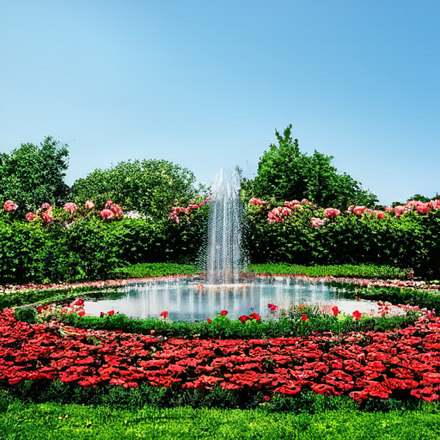} \\
    \includegraphics[width=.19\linewidth]{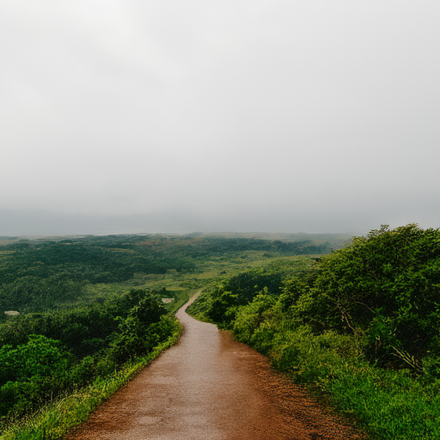} & \includegraphics[width=.19\linewidth]{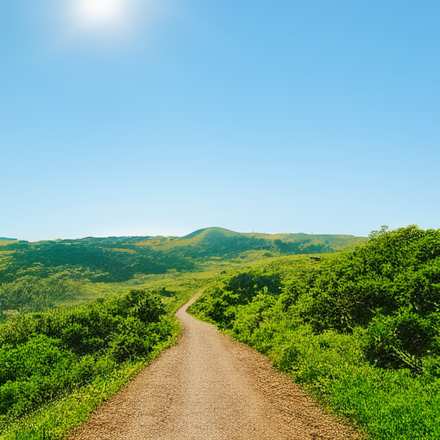} &
    \includegraphics[width=.19\linewidth]{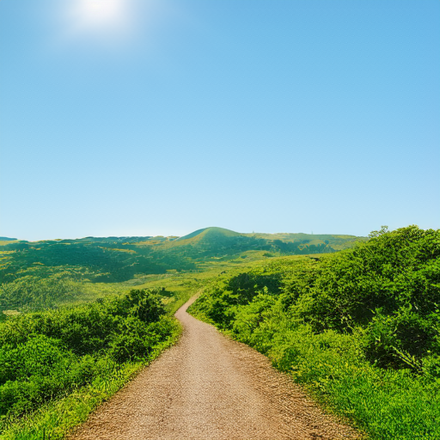} &    \includegraphics[width=.19\linewidth]{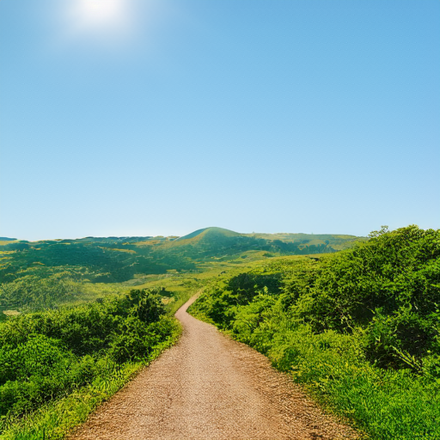} \\
    \includegraphics[width=.19\linewidth]{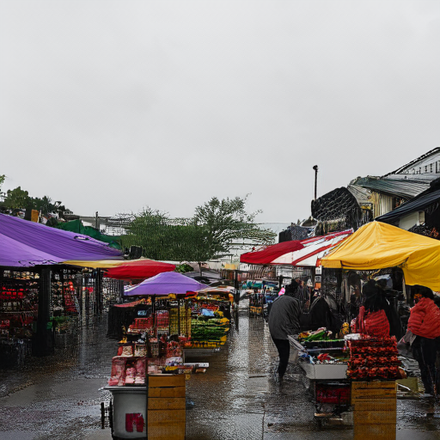} & \includegraphics[width=.19\linewidth]{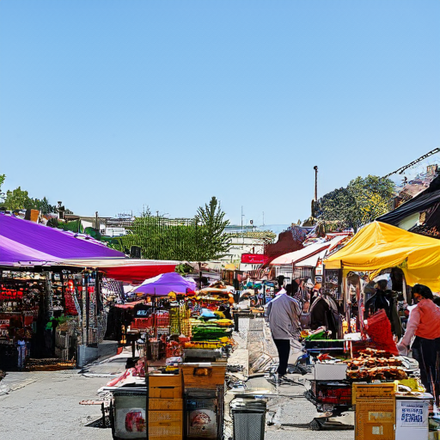} &
    \includegraphics[width=.19\linewidth]{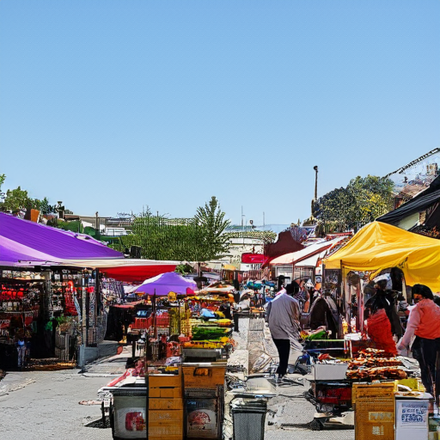} &    \includegraphics[width=.19\linewidth]{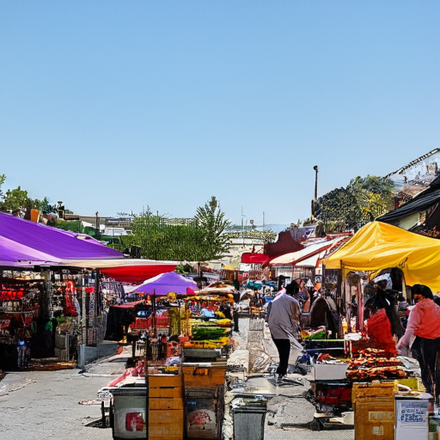} \\
    & \multicolumn{3}{c}{\emph{make it sunny}} \\
     \includegraphics[width=.19\linewidth]{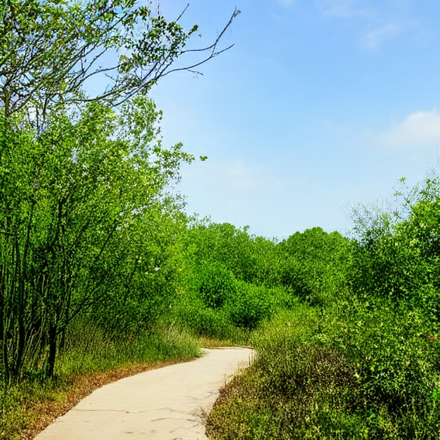} & \includegraphics[width=.19\linewidth]{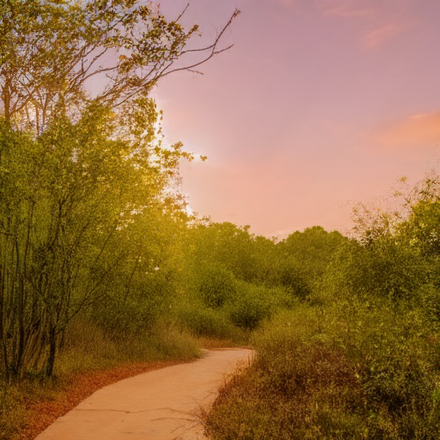} &
    \includegraphics[width=.19\linewidth]{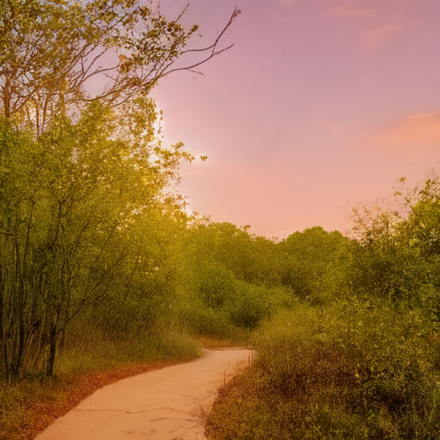} &    \includegraphics[width=.19\linewidth]{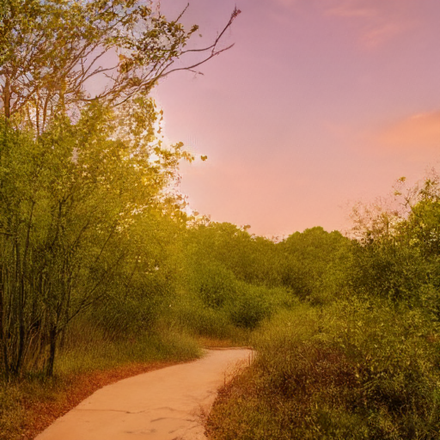} \\
    \includegraphics[width=.19\linewidth]{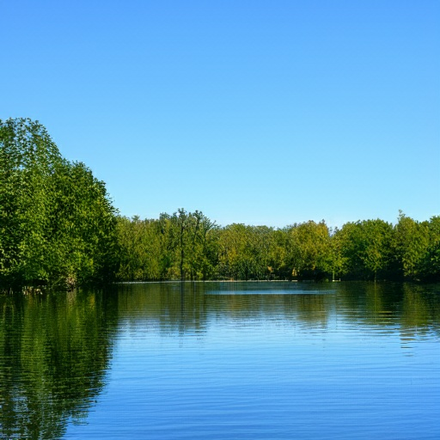} & \includegraphics[width=.19\linewidth]{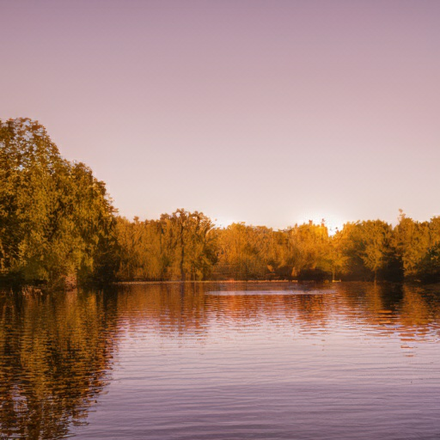} &
    \includegraphics[width=.19\linewidth]{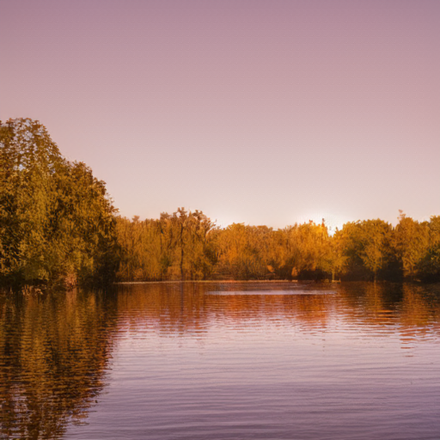} &    \includegraphics[width=.19\linewidth]{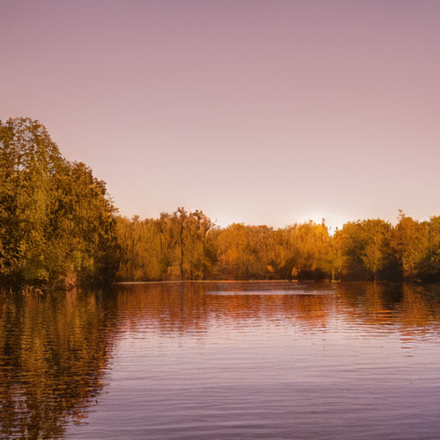} \\
    \includegraphics[width=.19\linewidth]{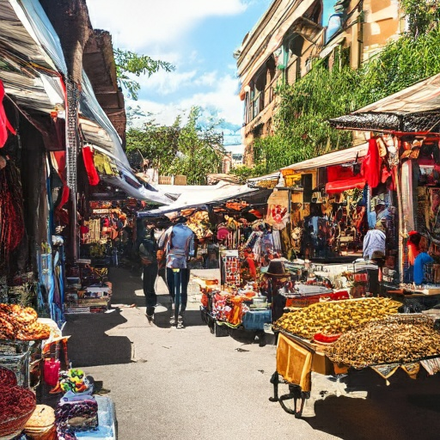} & \includegraphics[width=.19\linewidth]{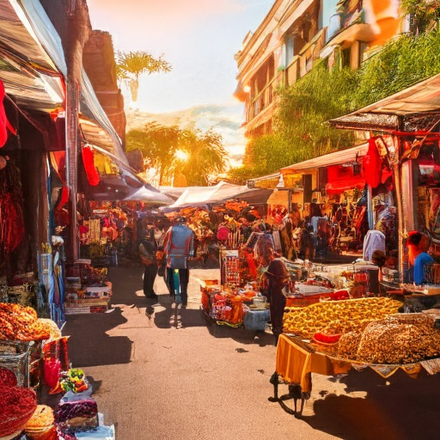} &
    \includegraphics[width=.19\linewidth]{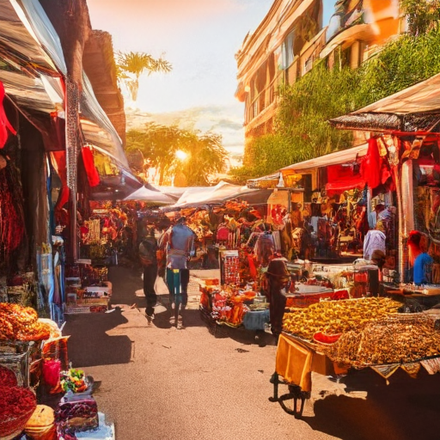} &    \includegraphics[width=.19\linewidth]{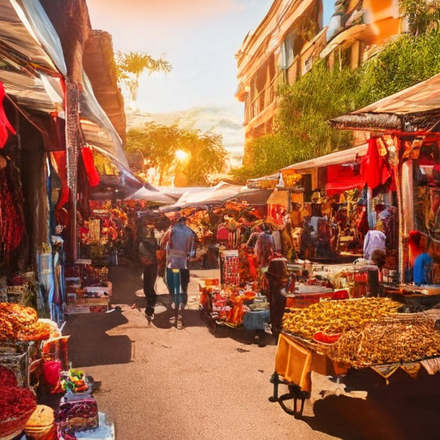} \\
    & \multicolumn{3}{c}{\emph{make it sunset}} \\
    \end{tabular}
    \caption{Visual comparisons with the IP2P model and our conditional distilled model.}
\end{figure}

\begin{figure}[htbp]
    \centering
    \setlength{\tabcolsep}{4pt}
    \begin{tabular}{c | c c c}
    \small \textbf{Input} & \small \textbf{IP2P} (200 steps) & \small \textbf{Ours} (1 step) & \small \textbf{Ours} (4 step) \\
    \includegraphics[width=.19\linewidth]{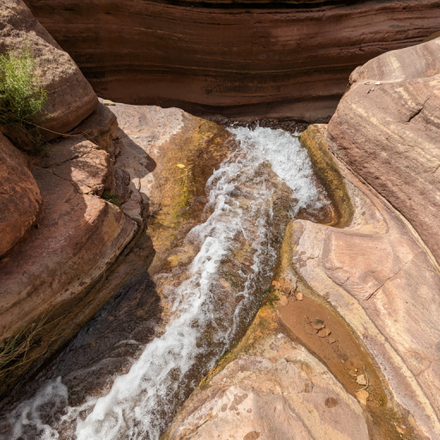} & \includegraphics[width=.19\linewidth]{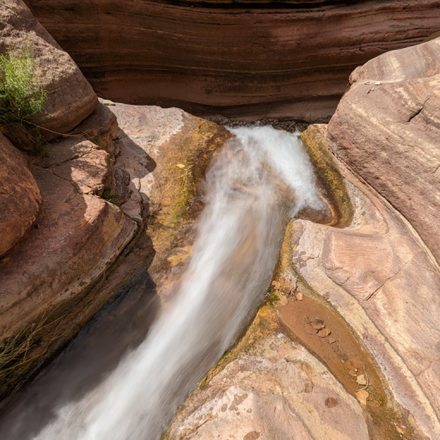} &
    \includegraphics[width=.19\linewidth]{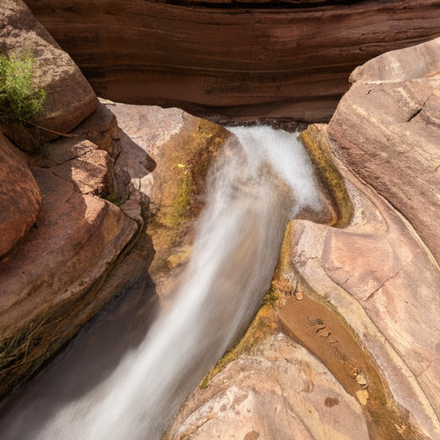} &    \includegraphics[width=.19\linewidth]{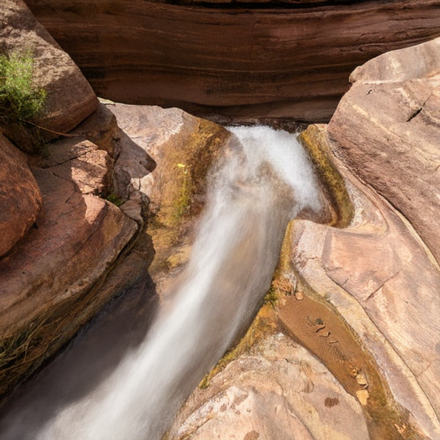} \\
    \includegraphics[width=.19\linewidth]{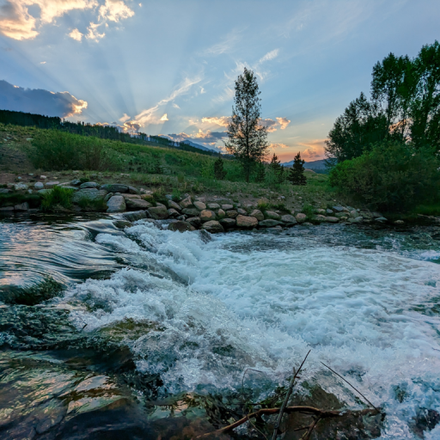} & \includegraphics[width=.19\linewidth]{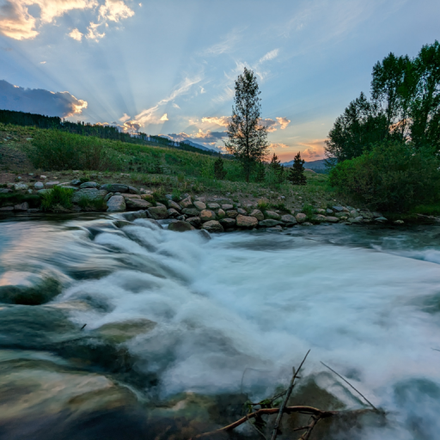} &
    \includegraphics[width=.19\linewidth]{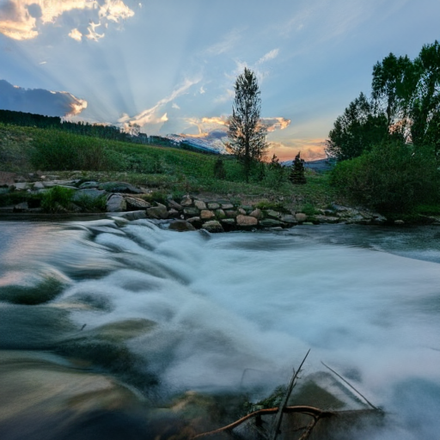} &    \includegraphics[width=.19\linewidth]{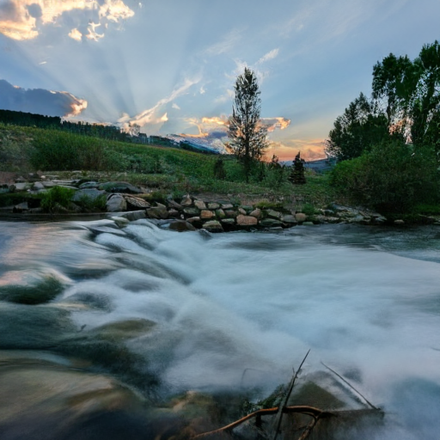} \\
    \includegraphics[width=.19\linewidth]{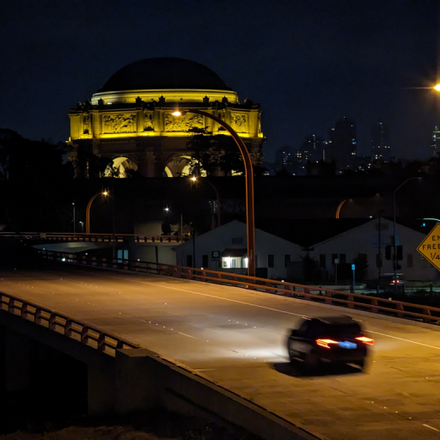} & \includegraphics[width=.19\linewidth]{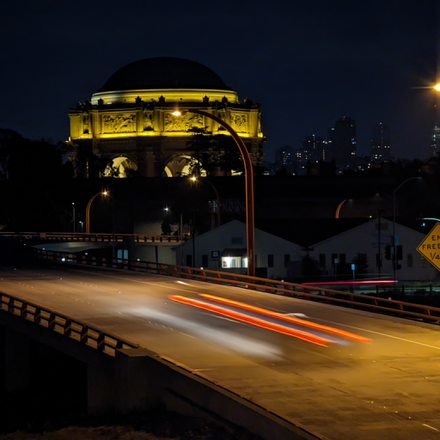} &
    \includegraphics[width=.19\linewidth]{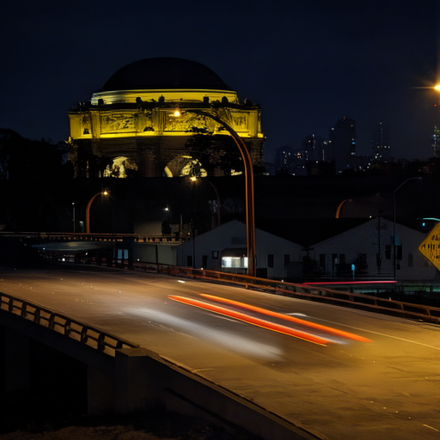} &    \includegraphics[width=.19\linewidth]{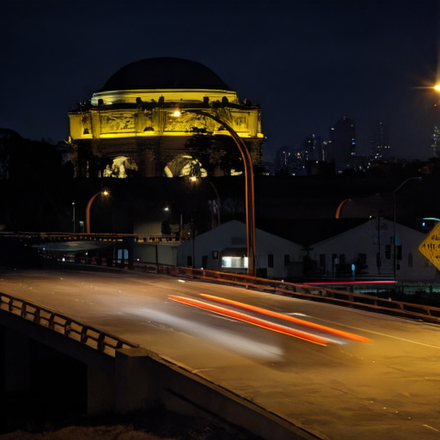} \\
    & \multicolumn{3}{c}{\emph{make it long exposure}} \\
    \includegraphics[width=.19\linewidth]{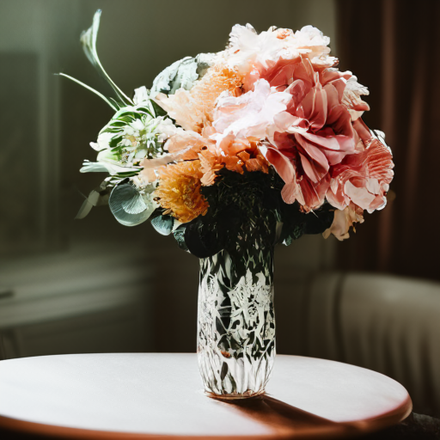} & \includegraphics[width=.19\linewidth]{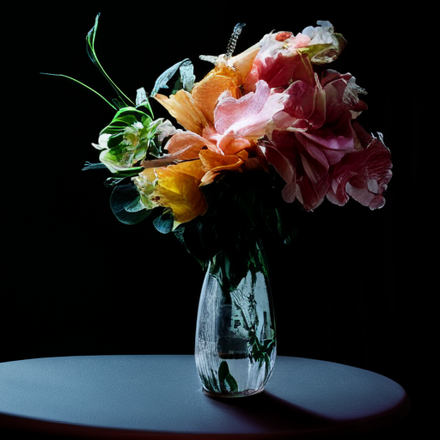} &
    \includegraphics[width=.19\linewidth]{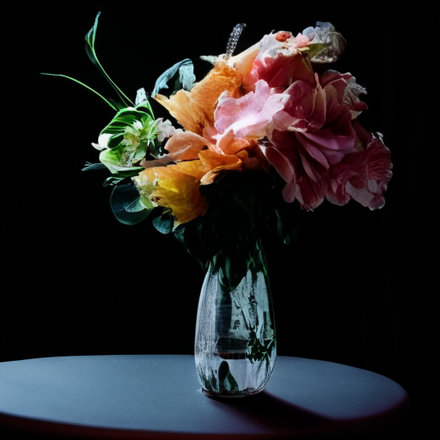} &    \includegraphics[width=.19\linewidth]{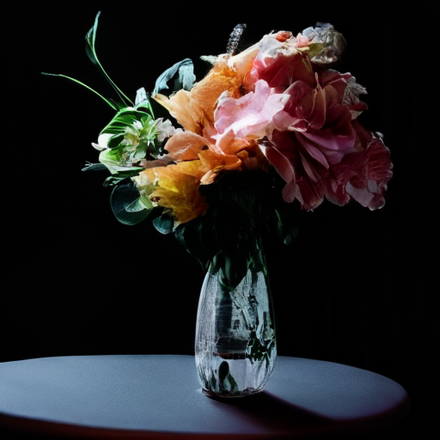} \\
    \includegraphics[width=.19\linewidth]{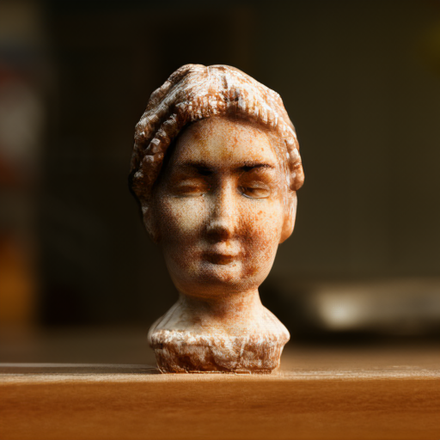} & \includegraphics[width=.19\linewidth]{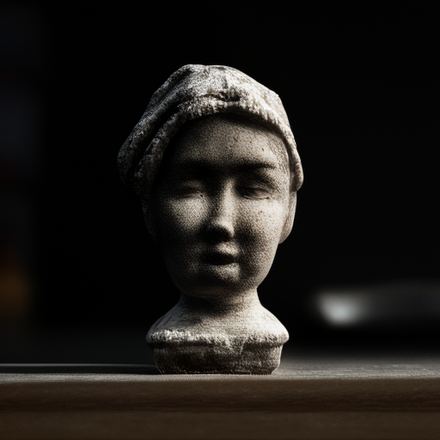} &
    \includegraphics[width=.19\linewidth]{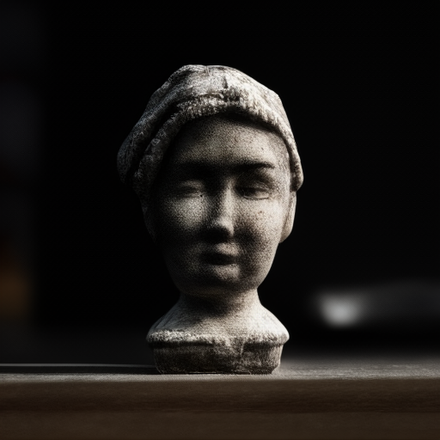} &    \includegraphics[width=.19\linewidth]{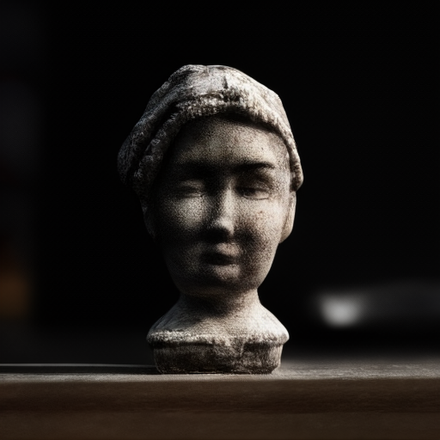} \\
     \includegraphics[width=.19\linewidth]{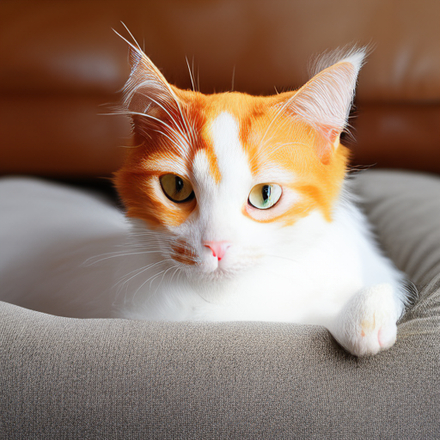} & \includegraphics[width=.19\linewidth]{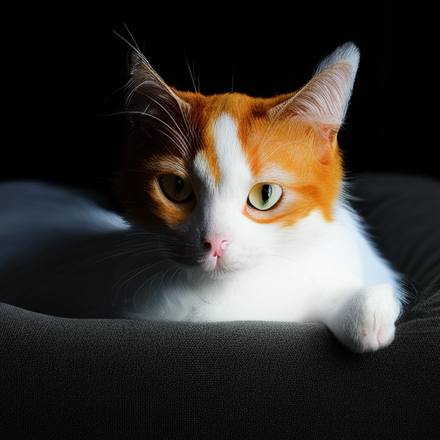} &
    \includegraphics[width=.19\linewidth]{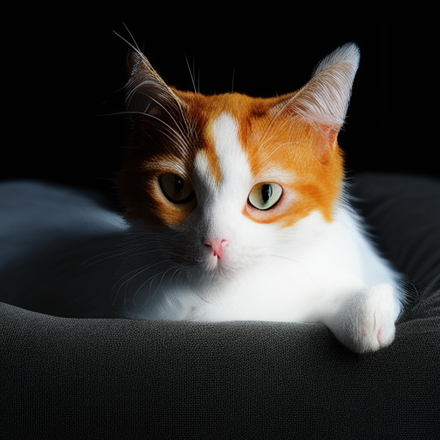} &    \includegraphics[width=.19\linewidth]{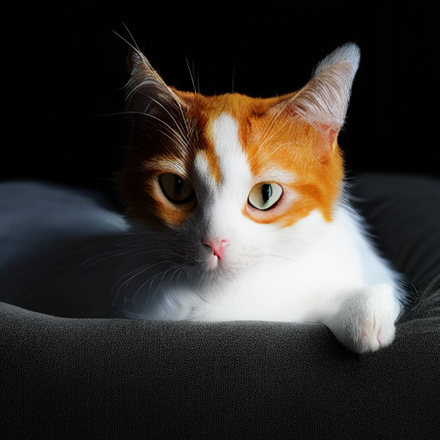} \\
    & \multicolumn{3}{c}{\emph{make it lowkey}} \\
    \end{tabular}
    \caption{Visual comparisons with the IP2P model and our conditional distilled model.}
\end{figure}